\documentclass[10pt,journal]{IEEEtran}

\usepackage[pdftex]{graphicx}
\DeclareGraphicsExtensions{.pdf,.jpeg,.png}

\usepackage{multirow,setspace,verbatim,amsfonts,graphicx,amsmath,amsthm,amsbsy,amssymb,epsfig,url,cite}
\usepackage{booktabs}

\usepackage{amsthm}
\usepackage{gensymb}
\usepackage{graphicx}
\usepackage{amsfonts}
\usepackage{amsmath,bbm}
\usepackage{amssymb}
\usepackage{amsthm}
\usepackage{tikz,graphicx}
\usepackage{mathrsfs}
\usepackage{algpseudocode}

\usepackage{algorithmicx}

\usepackage{array}
\usepackage{booktabs}

\newcommand{\omegab}{{\boldsymbol {\omega}}}

\newcommand{\Rd}{{\mathbb R}}

\newcommand{\Pc}{{{\mathcal P}}}

\newcommand{\rank}{\textsc{Rank}}
\newcommand{\hank}{\mathbb{H}}

\renewcommand{\vec}{\textsc{Vec}}

\renewcommand{\vec}{\textsc{Vec}}

\begin{document}

\title{Efficient B-mode Ultrasound Image Reconstruction from Sub-sampled RF Data using Deep Learning}

\author{Yeo Hun Yoon,~
Shujaat Khan,
        Jaeyoung Huh,~
        and~Jong~Chul~Ye,~\IEEEmembership{Senior Member,~IEEE}
\thanks{Copyright (c) 2017 IEEE. Personal use of this material is permitted. However, permission to use this material for any other purposes must be obtained from the IEEE by sending a request to pubs-permissions@ieee.org.}
\thanks{The authors are with the Department of Bio and Brain Engineering, Korea Advanced Institute of Science and Technology (KAIST), 
		Daejeon 34141, Republic of Korea (e-mail:\{caffemocha,shujaat,woori93,jong.ye\}@kaist.ac.kr).  {Parts of this work was presented in  2018 IEEE International Conference on Acoustic, Speech and Signal Processing (ICASSP)
		\cite{yoon2017deep}. Code and dataset is available at https://github.com/BISPL-JYH/Ultrasound\_TMI.}}
}

\maketitle

\begin{abstract}
In portable, three dimensional, and ultra-fast ultrasound  imaging systems, there is an increasing demand for the reconstruction of high quality images from a limited number of radio-frequency (RF) measurements due to receiver (Rx) or transmit (Xmit) event  sub-sampling.  However, due to the presence of side lobe artifacts from RF sub-sampling, the standard beam-former often produces blurry images with less contrast, which are unsuitable for diagnostic purposes.  Existing compressed sensing approaches often require either hardware changes or computationally expensive algorithms, but their quality improvements are limited.  To address this problem, here we propose a novel deep learning approach that directly interpolates the missing RF data by utilizing redundancy in the Rx-Xmit plane.  Our extensive experimental results using sub-sampled RF data from a multi-line acquisition B-mode system confirm that the proposed method can effectively reduce the data rate without sacrificing image quality.  
\end{abstract}


\IEEEpeerreviewmaketitle

\section{Introduction}
\label{sec:introduction}

\IEEEPARstart{D}{ue} to the excellent temporal resolution with reasonable image quality and minimal invasiveness, ultrasound (US) imaging has been adopted as a golden-standard for the diagnosis of many diseases in the heart, liver, etc.  Accordingly, there have been many research efforts to extend US imaging to new applications such as portable imaging in emergency care \cite{nelson2008use}, 3-D imaging \cite{lockwood1998real}, ultra-fast imaging \cite{mace2011functional,tanter2014ultrafast}, etc.

To achieve better spatial resolution in US imaging, high-speed analog-to-digital converters (ADC) should be used for the Rx portion of the US transducer, which consumes significant power.  Accordingly, in portable US systems, a small number of Rx elements with reduced aperture sizes are used to reduce power consumption, which often results in the deterioration of image quality.  On the other hand, to achieve higher frame rates, the number of transmit events should be reduced, as the duration of the transmit event is determined by the speed of sound.  This in turn leads to down-sampling artifacts.

To address these problems, compressed sensing (CS) approaches have been investigated \cite{liebgott2012pre,lorintiu2015compressed,wagner2012compressed,quinsac20103d,schretter2018ultrasound}.  However, US specific properties often deteriorate the performance of CS approaches.  For example, due to the wave nature of ultrasound scattering, it is often difficult to accurately model the sensing matrix.  Moreover, US images contain characteristic speckles, which make them barely sparse in any basis.  Instead of using wave scattering physics, Wagner \textit{et al}. \cite{wagner2012compressed} modeled a scan line profile as a signal with a finite rate of innovations (FRI) \cite{vetterli2002sampling}, and proposed a specially designed hardware architecture that enables high-resolution scan line reconstruction \cite{wagner2012compressed}.  One of the drawbacks of this approach, however, is that it cannot be used for conventional B-mode imaging systems.

Recently, inspired by the tremendous success of deep learning in classification \cite{krizhevsky2012imagenet,he2016deep,ganin2016domain} and low-level computer vision problems \cite{ronneberger2015u,zhang2016beyond,kim2016accurate}, many researchers have investigated deep learning approaches  for various biomedical image reconstruction problems and successfully demonstrated significant performance improvements over CS approaches \cite{kang2017deep,chen2017low,kang2018deep,chen2017lowBOE,adler2018learned,wolterink2017generative,jin2017deep,han2017framing,wang2016accelerating,hammernik2018learning,schlemper2018deep,zhu2018image,lee2018deep}.  The source of deep learning success in image reconstruction lies in its exponentially increasing expressiveness, which can capture modality-specific image features \cite{lecun2015deep}.  Therefore, this paper aims at developing a deep learning algorithm that provides efficient image reconstruction from sub-sampled RF data.

In US literature, the works in \cite{Allman_reviewer,Luchies_reviewer2} were among the first to apply deep learning approaches to US image reconstruction.  In particular, Allman \textit{et al} \cite{Allman_reviewer} proposed a machine learning method to identify and remove reflection artifacts in photoacoustic channel data.  Luchies and Byram \cite{Luchies_reviewer2} proposed an ultrasound beam-forming approach using deep neural networks.  These works were more focused on processing fully-sampled channel RF data than reconstruction from sub-sampled channel RFs.  To the best of our knowledge, no existing deep learning approaches address image recovery from channel RF sub-sampled data.

One of the most important contributions of this work is that it demonstrates that deep neural networks can estimate missing RF data from Rx and/or Xmit event subsampling without sacrificing image quality.  However, a deep neural network is usually considered a black box, so its application to medical image reconstruction often causes skepticism as to whether the enhancement is cosmetic or real.  Thus, in contrast to common practice, our deep neural network is carefully designed based on theoretical justification.  More specifically, our earlier work \cite{jin2016compressive} showed that strong correlation in RF data results in a rank deficient Hankel structured matrix so that missing RF data can be estimated using an annihilating filter-based low rank Hankel matrix approach (ALOHA) \cite{jin2016general,lee2016acceleration,lee2016reference,ye2016compressive,haldar2014low}.  In another of our group's previous works \cite{ye2017deep}, we discovered that deep neural networks are closely related to low-rank Hankel matrix decomposition using data-driven convolutional framelets.  By synergistically combining these findings, we propose a deep neural network that performs direct RF data interpolation by exploiting RF data redundancy in a fully data-driven manner.

In addition to the theoretical justification, there are several benefits from our approach.  First, in contrast to the low-rank matrix completion approach for RF interpolation \cite{jin2016compressive}, our deep network has a run-time complexity that is several orders of magnitude lower.  However, it is important to note that the proposed neural network is not another implementation of ALOHA for computational saving; rather, it is a new algorithm that significantly improves the interpolation performance of ALOHA by exploiting the exponential expressiveness of a deep neural network as confirmed later by extensive experimental results.

Another important benefit to exploiting the link between ALOHA and the deep neural network is that it guides us to choose an appropriate domain to implement a deep neural network.  Since ALOHA successfully interpolated RF data in our prior work \cite{jin2016compressive}, our theoretical understanding of the link led us to implement a neural network in the RF domain, which may be unexpected from the perspective of implementing a standard deep neural network.  Compared to image domain CNNs that attempt to learn acquisition geometry specific artifacts, one of the most important advantages of the proposed RF domain CNN is its generalization power.  For example, although an image domain deep learning approach requires many sets of data from different acquisition geometries and body areas \cite{kang2017deep}, our CNN can be trained using RF data  measured by a specific transducer for a particular organ, but it can also be used for other types of transducers and/or different organs.

The rest of the paper is organized as follows.  Section~\ref{sec:physics} reviews the B-mode US acquisition physics.  In Section~\ref{sec:theory}, we explain the inherent redundancy in RF data, and how it can be exploited using a deep neural network.  Experimental methods and results are presented in Section~\ref{sec:methods} and Section~\ref{sec:results}, which is followed by the conclusions in Section~\ref{sec:conclusion}.

\section{Imaging Physics}
\label{sec:physics}

\subsection{B-mode Ultrasound Imaging}

\begin{figure}[!hbt]
	\begin{minipage}[b]{0.45\linewidth}
		\centering
		\centerline{\includegraphics[width=2.5cm,height=2.5cm]{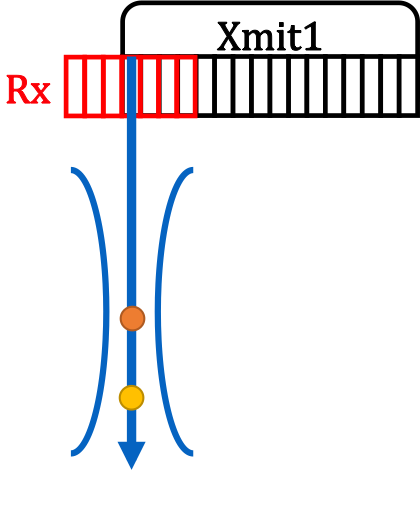}}
		\centerline{(a) Beam emission}\medskip
	\end{minipage}
	\hfill
	\begin{minipage}[b]{0.45\linewidth}
		\centering
		\centerline{\includegraphics[width=3.5cm,height=3.cm]{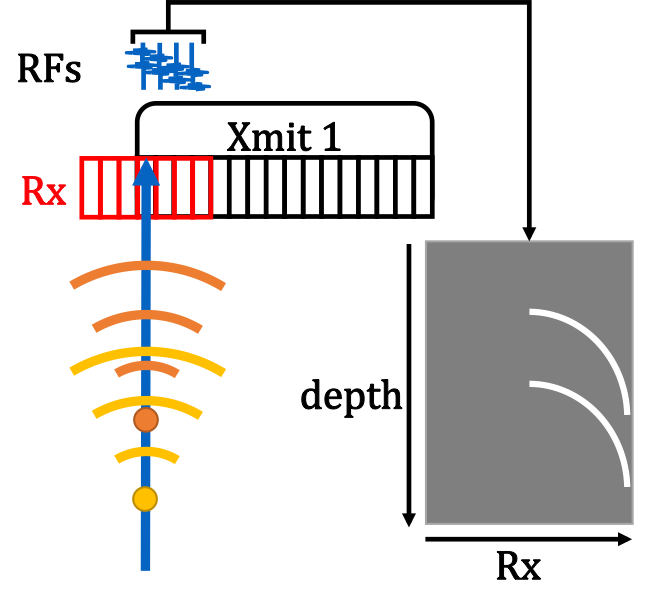}}
		\centerline{(b) RF data recording}\medskip
	\end{minipage}
	\begin{minipage}[b]{0.90\linewidth}
		\centering
		\centerline{\includegraphics[width=8cm]{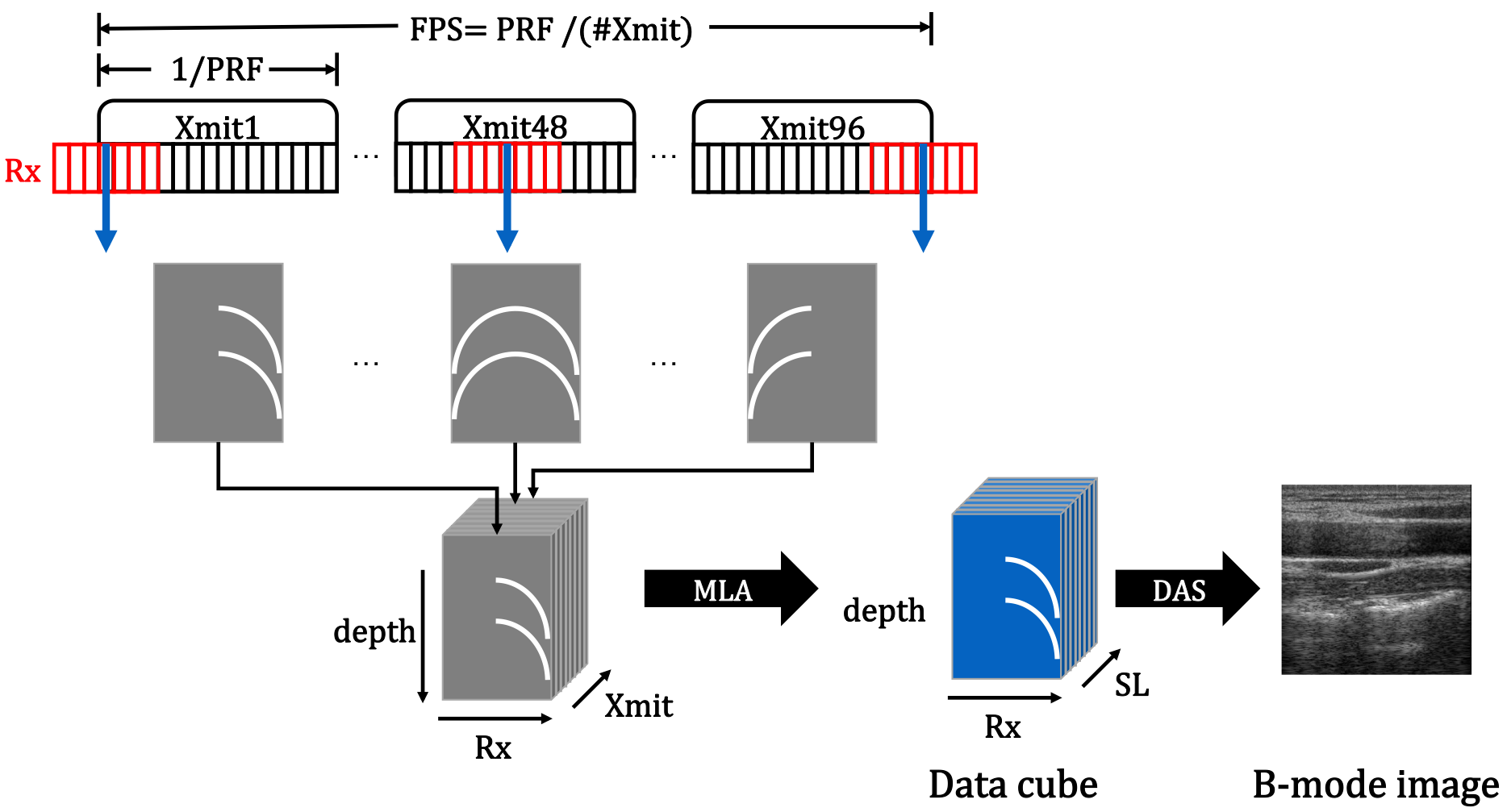}}
		\centerline{(c) DAS beamforming}
	\end{minipage}
	\caption{Imaging flow of the standard B-mode ultrasound imaging. PRF : pulse repetition frequency, FPS : frame per second.}
	\label{fig:theorya}
\end{figure}

 B-mode ultrasound imaging, which is most widely used in practice, scans the body using focused beams and reconstruct 2-D images as shown in Fig.~\ref{fig:theorya}.  Here, Xmit, Rx, and DAS denote the transmit event for each ultrasound beam, the receivers of the transducer, and the delay-and-sum beam-former, respectively. Specifically, after a focused ultrasound beam is transmitted as shown in Fig.~\ref{fig:theorya}(a), the ultrasound beam is reflected from some tissue boundaries and the reflected US data is recorded by Rx as RF data (see Fig.~\ref{fig:theorya}(b)).  Thus, Depth-Rx coordinate RF data is obtained for each transmit event, and this is repeated to obtain a 3-D cube of RF data in the Depth-Rx-Xmit coordinates.  

\begin{figure*}[!t]
	\centering
		\centerline{\includegraphics[width=0.8\linewidth]{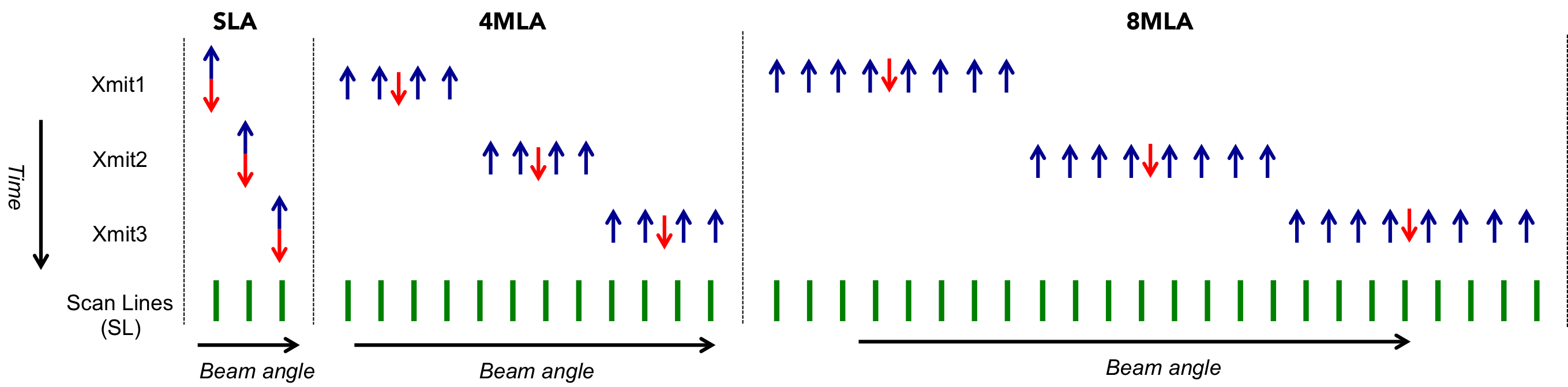}}
	\caption{Comparison of SLA, 4MLA, and 8MLA \cite{misaridis2001ultrasound}.   Down arrows indicate transmit directions and up arrows receive directions. For SLA, transmit and receive directions are identical.  For MLA, different scan lines are generated by shifting Rx measurements. 	The synthesized scan line directions are indicated with lines at the bottom of the illustration.}
	\label{fig:MLA}
\end{figure*}

In practice, this B-mode acquisition is usually combined with multi-line acquisition (MLA) or parallel beam-forming (multiple parallel receive beams for each transmit beam) \cite{shattuck1984explososcan,von1991high,misaridis2001ultrasound} to increase the frame rate.  The concept is illustrated in Fig.~\ref{fig:MLA}, which shows the MLA procedure for four and eight parallel beams (denoted by 4MLA and 8MLA, respectively).  More specifically, for each transmit event, additional scan lines (SLs) are synthesized from the same Rx measurements using pre-defined offsets and weights \cite{shattuck1984explososcan,von1991high,misaridis2001ultrasound}.  This can increase the frame-rate thanks to fewer transmit events with the same number of scan lines.

Finally, a DAS beam-former uses the resulting Depth-Rx-SL data to generate an image frame (see Fig.~\ref{fig:theorya}(c)).  Depending on the scanner, the DAS beam-former can be implemented using either hardware or software.

\subsection{RF Sub-sampled MLA}

\begin{figure}[!h]
	\centering
		\centerline{\includegraphics[width=0.9\linewidth]{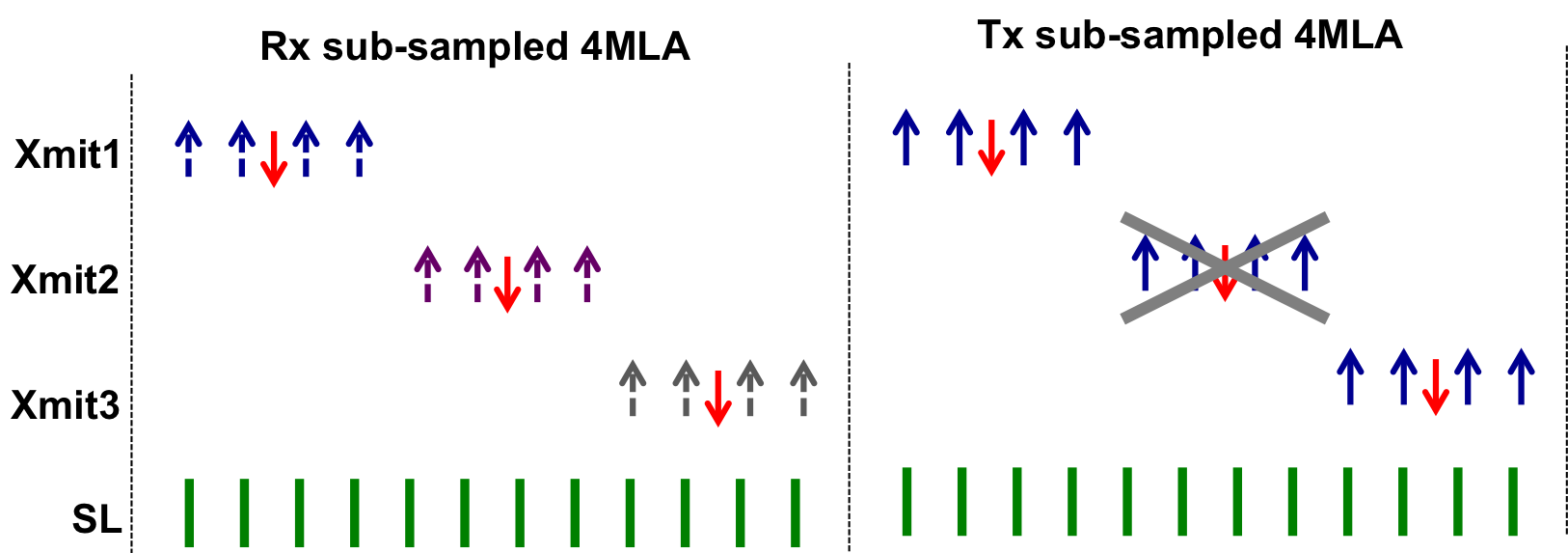}}
\centerline{\mbox{(a) \hspace{4cm}\mbox{(b)}}}
				\vspace*{0.3cm}
	\hspace*{0.1cm}\centerline{\includegraphics[width=4.5cm]{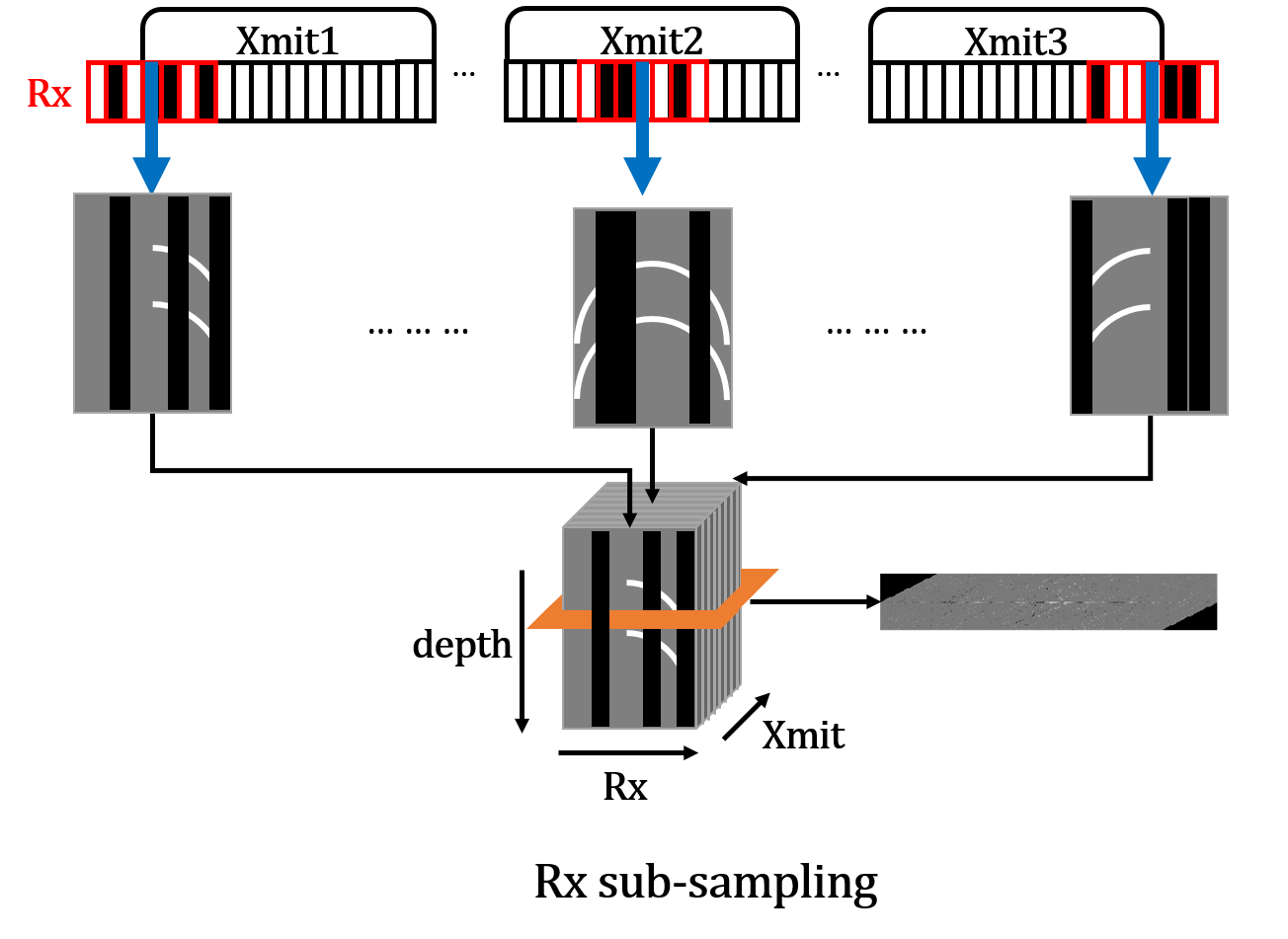}\hspace*{0.1cm}\includegraphics[width=4.5cm]{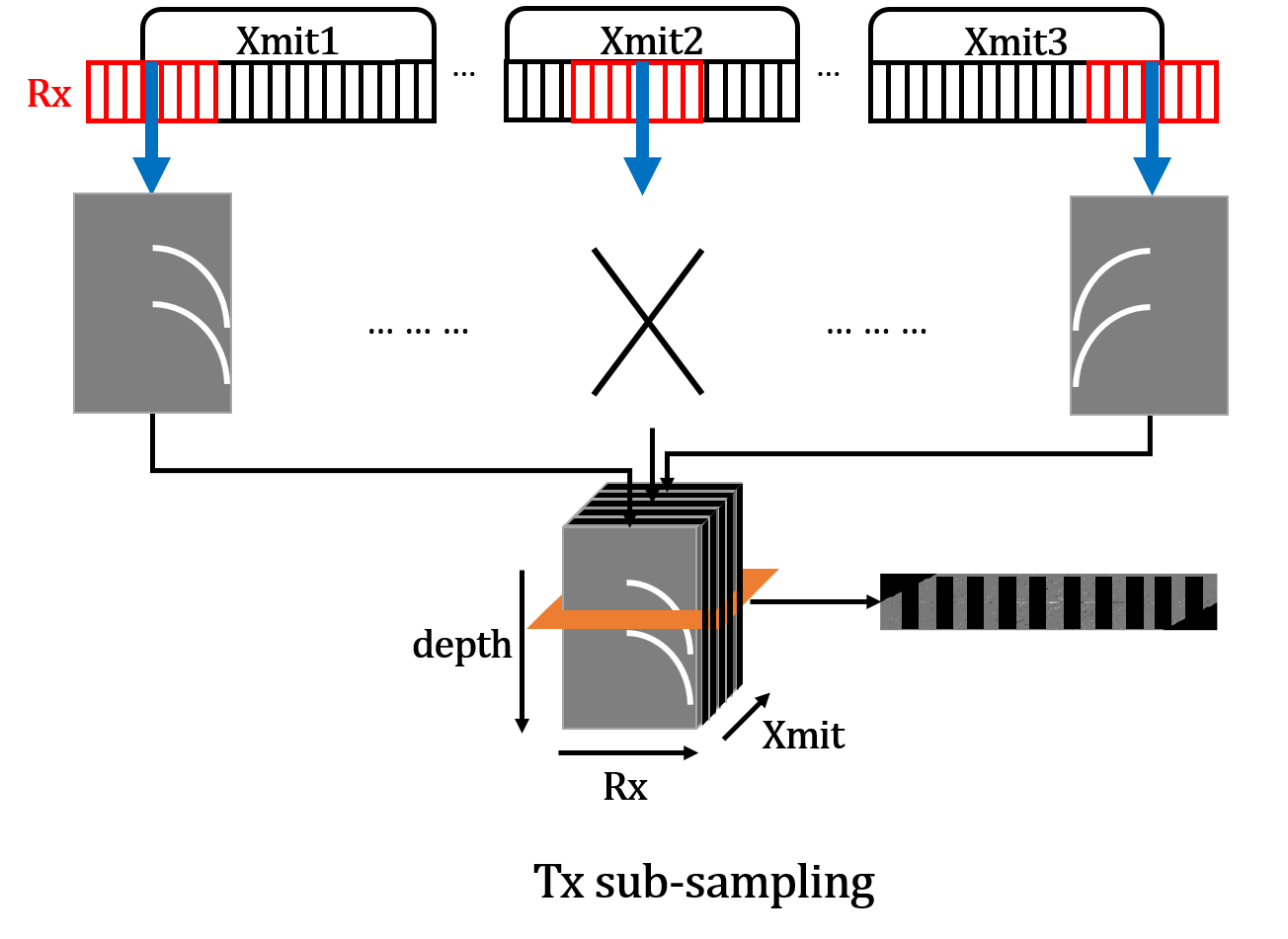}}	
	\centerline{\mbox{(c) \hspace{4cm}\mbox{(d)}}}

	\caption{RF subsampled MLA:  (a)(c) Rx sub-sampled MLA  and its acquisition example. Same color-coded dashed up arrows indicates the receiver direction that are affected from the same Rx subsampling. (b)(d) Tx sub-sampled MLA and its acquisition example.  Once a transmit event is skipped, all the scan lines synthesized from the transmit cannot be generated. Down arrows indicate transmit directions and up arrows receive directions. The synthesized scan line directions are indicated with lines at the bottom of the illustration.
}
		\vspace*{-0.3cm}
	\label{fig:MLA-sub}
\end{figure}

Here, we consider two types of RF sub-sampling schemes.  The first is a random Rx sub-sampling \cite{jin2016compressive} that acquires only random subsets of Rx data for each transmit event.  When combined with MLA, the same sub-sampled Rx data affect multiple parallel beams for each transmit event as shown in Figs.~\ref{fig:MLA-sub}(a)(c).  This sub-sampling scheme can be easily used to reduce data transmission bandwidth and power-consumption.  However, it is usually associated with image artifacts and contrast loss.  Second, we consider uniform sub-sampling of the Xmit event as shown in Figs.~\ref{fig:MLA-sub}(b)(d) to further increase the frame rate.  However, once a transmit event is skipped, all the associated SLs from the transmit event cannot be synthesized as shown in Figs.~\ref{fig:MLA-sub}(b)(d).  To retain the same number of scan lines, the number of parallel beams should be increased accordingly.  However, use of the same Rx measurement to generate a large number of SLs introduces image artifacts and reduces the contrast.

\section{Main Contribution}
\label{sec:theory}

Since these RF-subsampling schemes are usually associated with artifacts, in this section, a neural network is designed to achieve high quality reconstruction from the RF sub-sampling schemes. 

\subsection{Redundancy of RF data and low-rank Hankel matrix}

\begin{figure}[!htb]
  \centering
  \centerline{\includegraphics[width=7cm]{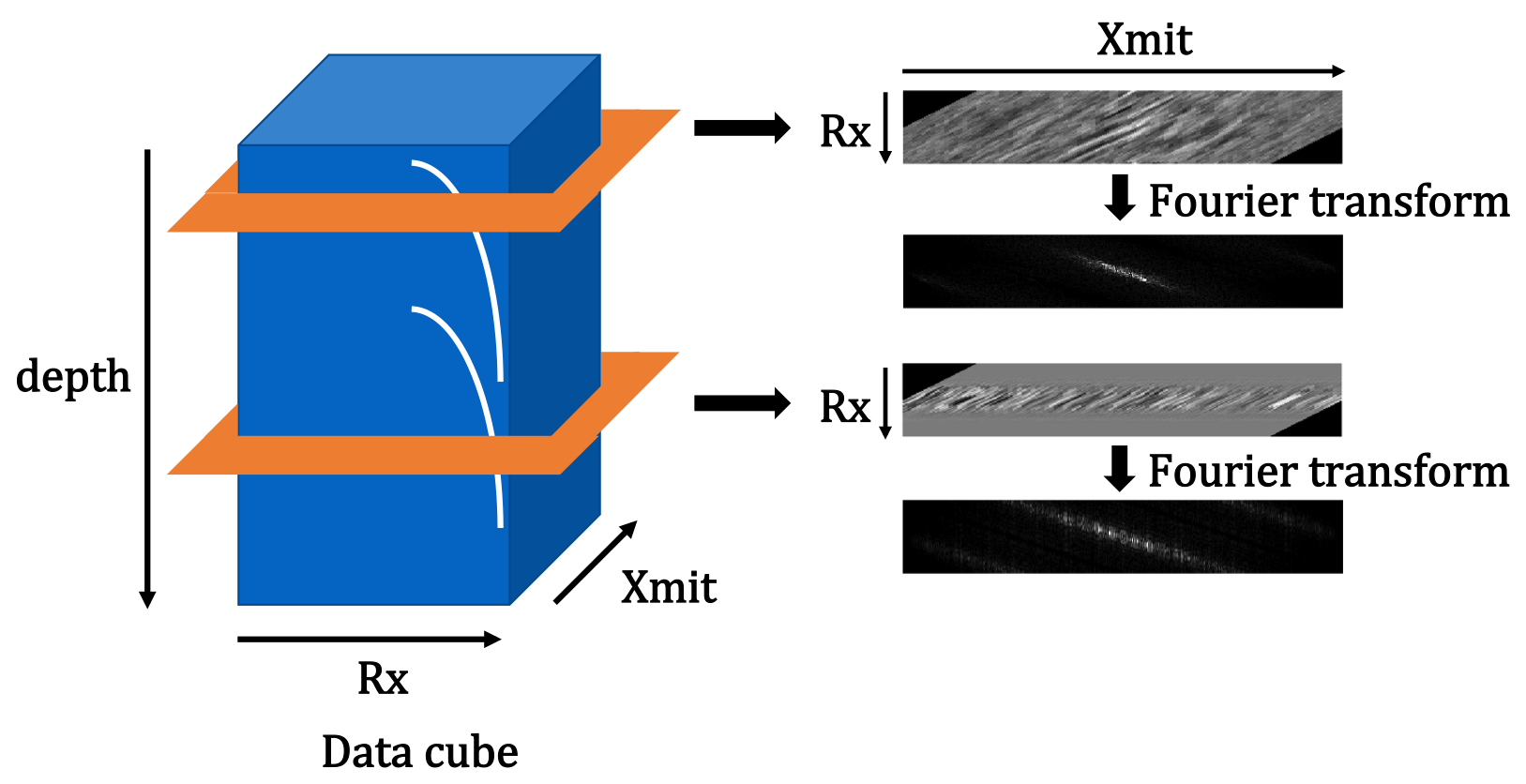}}
\caption{Rx-Xmit data from depth-Rx-Xmit data cube and its Fourier spectrum.}
\label{fig:methoda}
\end{figure}

In B-mode ultrasound imaging, the direction of each transmit event changes incrementally, so the acquired Rx data does not change rapidly for each transmit event.  This implies that some degree of skew redundancy exists in the Rx-Xmit (or Rx-SL) coordinate data and it can be easily seen as sparsity in the Fourier domain as demonstrated in Fig.~\ref{fig:methoda}.

 Therefore, when the Fourier spectrum of the Rx-Xmit image is denoted by $\widehat{F}(\omegab)$, we can find a function $\widehat{K}(\omegab)$ in the spectral domain such that their multiplication becomes zero \cite{jin2016general,lee2016acceleration,lee2016reference,ye2016compressive}:
\begin{equation}
\widehat{F}(\omegab) \widehat{K}(\omegab) =0  , \quad \forall \omegab \quad .
\end{equation}
This is equivalent to the convolutional relationship in the Rx-Xmit domain:
\begin{eqnarray}
{F} \circledast {K} =0,\quad \quad 
\end{eqnarray}
where $F \in \Rd^{n_1\times n_2}$ denotes the discrete 2-D image and
$K \in \Rd^{d_1\times d_2}$ is often referred to as the {\em annihilating filter} \cite{vetterli2002sampling}.
This can be represented in a matrix form:
\begin{eqnarray}\label{eq:FK}
\hank_{d_1,d_2}(F) \overline{\vec(K)} =0,
\end{eqnarray}
where $\vec(K)$ denotes the vectorization operation by stacking the column vectors of the 2-D matrix $K$, and $\overline{\vec(K)}$ is a flipped  (i.e. index reversed) version of the vector ${\vec(K)}$.
Here, 
$\hank_{d_1,d_2}(F) \in \Rd^{n_1n_2\times d_1d_2}$ is the
{\em block} Hankel matrix for the image $F=[f_1,\cdots, f_{n_2}]\in \Rd^{n_1\times n_2}$, which is defined under the periodic boundary condition as follows \cite{jin2016general,lee2016acceleration,lee2016reference,ye2016compressive}:
 \begin{eqnarray} \label{eq:Hank}
\hank_{d_1,d_2}(F) =\left[
        \begin{array}{cccc}
        \hank_{d_1}(f_1)  &   \hank_{d_1}(f_2)& \cdots   &   \hank_{d_1}(f_{d_2})   \\
      \hank_{d_1}(f_2) &  \hank_{d_1}(f_3) & \cdots &     \hank_{d_1}(f_{d_2+1}) \\
           \vdots    & \vdots     &  \ddots    & \vdots    \\
          \hank_{d_1}(f_{n_2}) &   \hank_{d_1}(f_1)& \cdots &   \hank_{d_1}(f_{d_2-1})  \\
        \end{array}
    \right],
    \end{eqnarray}
 and
 $\hank_{d_1}(f_i)\in \Rd^{n_1\times d_1}$ is a Hankel matrix:
 \begin{eqnarray} \label{eq:hank}
\hank_d(f_i) =\left[
        \begin{array}{cccc}
        f_i[1]  &   f_i[2] & \cdots   &   f_i[d_1]   \\
       f_i[2]  &   f_i[3] & \cdots &     f_i[d_1+1] \\
           \vdots    & \vdots     &  \ddots    & \vdots    \\
              f_i[n_1]  &   f_i[1] & \cdots &   f_i[d_1-1] \\
        \end{array}
    \right].
    \end{eqnarray}

Eq.~\eqref{eq:FK} implies that the block Hankel matrix constructed from the RF data in the Rx-Xmit domain is rank-deficient. Furthermore, its rank is determined by the sparsity level in the spectral domain as theoretically proven in \cite{ye2016compressive}. In fact, Jin \textit{et al}. \cite{jin2016compressive} utilized this to interpolate missing RF data using low-rank Hankel matrix completion by solving the following optimization problem:

\begin{eqnarray}
 (P) \quad & \min_{M\in \Rd^{n_1\times n_2}}  & \left\|F-M\right\|^2 \label{eq:imgcost}  \\
&\mbox{subject to }  &\rank~ \hank_{d_1,d_2}\left(M\right) =s , \label{eq:ccost} \\
&& 
\Pc_\Lambda \left[F\right]= \Pc_\Lambda \left[M\right],\quad  \notag
\end{eqnarray}
where $\Lambda$ denotes the measured RF indices.  This low-rank constraint optimization problem can be solved by matrix factorization \cite{jin2015annihilating,jin2016general,lee2016acceleration,lee2016reference,jin2016mri}.  However, the main limitation of \cite{jin2016compressive} is its computational complexity (see Appendix A for the details).  Moreover, \cite{jin2016compressive} additionally requires exploitation of temporal domain redundancy, which often results in reduced temporal resolution.

\subsection{RF Interpolation using Deep Learning}

It was recently demonstrated in \cite{ye2017deep} that a {\em properly designed} deep neural network is the signal space manifestation of the factorization of a {\em high-dimensionally lifted} signal.  Here, we briefly review the underlying idea.

Specifically, for a given signal $F \in \Rd^{n_1\times n_2}$, consider its high dimensional {\em lifting} using a Hankel matrix $\hank_{d_1,d_2}(F) \in \Rd^{n_1n_2\times d_1d_2}$ in \eqref{eq:Hank}.
Now, consider matrices $\Phi,\tilde\Phi \in \Rd^{n_1n_2\times m}$ which satisfy
\begin{eqnarray}
\tilde\Phi \Phi^\top &=& \alpha I_{n_1n_2},\quad \alpha > 0   \label{eq:frame} ,
\end{eqnarray}
where $I_n$ denotes the $n\times n$ identity matrix. In \cite{ye2017deep}, the matrices $\Phi,\tilde\Phi$ were identified as  the pooling and unpooling, respectively.  In addition, consider the additional set of matrices $\Psi,\tilde\Psi \in \Rd^{d_1d_2\times s}$ satisfying
\begin{eqnarray}
\Psi\tilde\Psi^\top &=& P_{R(V)} \label{eq:projection},
\end{eqnarray}
where $P_{R(V)}$ denotes the projection matrix to the range space of $V$, and $V$ is the basis vector for the row subspace of $\hank_{d_1,d_2}(F)$.  In \cite{ye2017deep}, $\Psi,\tilde\Psi$ are identified as learnable convolution filters.

Using Eqs.~\eqref{eq:frame} and \eqref{eq:projection}, it is trivial to see that
\begin{eqnarray}\label{eq:PR}
\hank_{d_1,d_2}(F) &=& \frac{1}{\alpha}\tilde \Phi \Phi^\top \hank_{d_1,d_2}(F) \Psi \tilde\Psi^\top , \notag\\
&=& \tilde \Phi C \tilde\Psi^\top ,
\end{eqnarray}
where the so-called convolution framelet coefficients are defined as
\begin{eqnarray}\label{eq:C}
C:= \frac{1}{\alpha}\Phi^\top \hank_{d_1,d_2}(F) \Psi  \quad .
\end{eqnarray}
Thanks to the property of the Hankel matrix, \eqref{eq:C} can be equivalently represented using the convolution \cite{ye2017deep}:
\begin{eqnarray}
C &=&\Phi^\top ( F \circledast H) , \label{eq:enc}
\end{eqnarray}
where the multi-channel filter  $H:=H(\Psi/\alpha)$ can be obtained by arranging the elements of $\Psi$ after scaling by $1/\alpha$.
Next, the high dimensional representation \eqref{eq:PR} can be {\em unlifted} to the original signal space as follows \cite{ye2017deep}:
\begin{eqnarray}
F &=& (\tilde \Phi C) \circledast  G , \label{eq:dec}
\end{eqnarray}
where the  multi-channel filter  $G:=G(\tilde\Psi)$ can be obtained by rearranging the elements of $\tilde\Psi$.

It is important to note that the convolutions in \eqref{eq:enc} and \eqref{eq:dec} are exactly the same convolutions used for the existing convolutional neural network \cite{ye2017deep}.  In fact, the structure in \eqref{eq:enc} and \eqref{eq:dec} corresponds to the popular encoder-decoder network architecture, where the number of filter channels is determined by the number of columns of $\Psi$ and $\tilde \Psi$.  This confirms that an encoder-decoder network can be derived as the signal space manifestation of the Hankel matrix decomposition in a high-dimensional space.

Based on these findings, it is easy to see that a feasible solution for $(P)$ satisfies the encoder-decoder network structure:
$$M = (\tilde \Phi C)\circledast G,\quad C:= \Phi^\top (M\circledast H),$$
where the number of channels for the filters $H$ and $G$ is $s$.  The authors in \cite{ye2017deep} found that although the pooling and unpooling operations $\Phi$ and $\tilde\Phi$ can be pre-defined based on domain knowledge,
the filters should be learned from the training data.
Then,  for a given set of sub-sampled RF data  $Y^{(i)}:=P_\Lambda \left[F^{(i)}\right]$ and fully sampled data $F^{(i)}$ for $i=1,\cdots, n_t$, the
filter learning problem can be readily obtained from $(P)$ as follows:
\begin{eqnarray}\label{eq:HG}
\min_{H,G} \sum_{i=1}^{n_t} \left\| F^{(i)} - \mathcal{K}\left(Y^{(i)};H,G\right)  \right\|^2 , \quad 
\end{eqnarray}
where  the operator $\mathcal{K}: \Rd^{n_1\times n_2} \to \Rd^{n_1\times n_2}$ is defined as
 $$\mathcal{K}\left(Y^{(i)};H,G\right)  =  (\tilde \Phi C(Y^{(i)}))\circledast G, $$ 
 in terms of the mapping $C: \Rd^{n_1\times n_2} \to \Rd^{n_1\times n_2}$
 $$C(Y) = \Phi^\top (M\circledast H),\quad \forall Y \in \Rd^{n_1\times n_2}.$$
 At the inference stage, the interpolated RF data can be readily obtained from the sub-sampled RF data $\tilde Y$ using the learned filters:
 $$\tilde F = \mathcal{K}\left(\tilde Y;H,G\right) .$$

 The filter learning problem in \eqref{eq:HG} can be easily extended for an encoder-decoder network with ReLU \cite{ye2017deep,han2018kspace}.
 The role of the rectified linear unit (ReLU) then guides the learning procedure such that the output signal can be represented as the nonnegative (conic) combination of the learned convolution framelet basis \cite{han2018kspace}.  This so-called {\em conic coding} is popular for learning part-by-part representations of objects \cite{lee1997unsupervised,lee1999learning}, which constitute the key ingredients of nonnegative matrix factorization (NMF) \cite{lee1999learning,lee2001algorithms}.

Simple signal expansion using \eqref{eq:enc} and \eqref{eq:dec} with ReLU is so powerful that a deep encoder-decoder neural network architecture emerges from recursive application of this convolutional framelet expansion with ReLU to the framelet coefficients \eqref{eq:C}.  The cascaded encoder-decoder network with ReLU results in an exponentially increasing expressiveness \cite{lecun2015deep} that can capture US specific image features that would not have been possible with ALOHA.  This is the primary motivation for our interest in using a deep neural network as a generalization beyond the low-rank Hankel matrix approach.

\begin{figure}[!htb]
	\centerline{\includegraphics[width=0.35\textwidth]{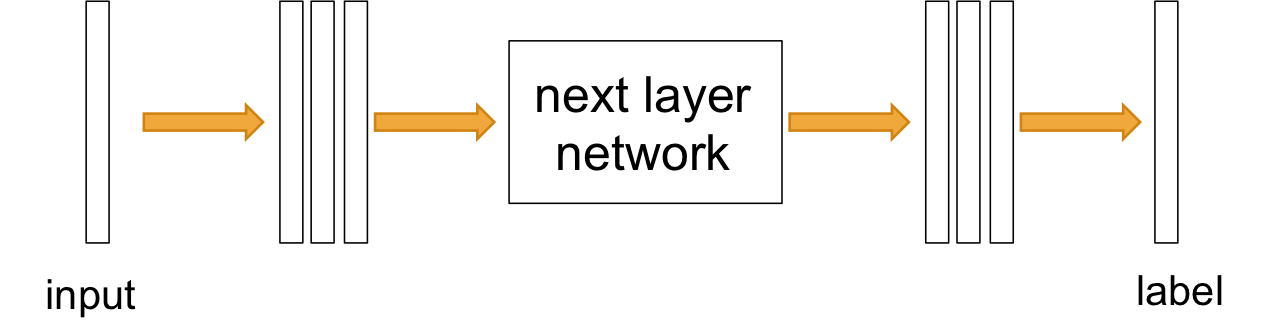}}
		\vspace*{-0.35cm}
	\centerline{\mbox{(a)}}
	\vspace*{0.35cm}
		\centerline{\includegraphics[width=0.35\textwidth]{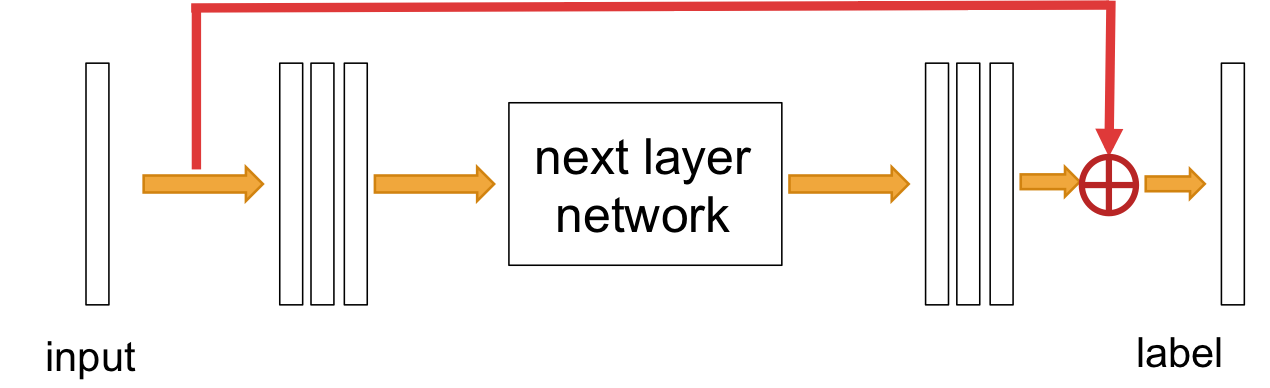}}
				\vspace*{-0.35cm}
			\centerline{\mbox{(b)}}

	\caption{Deep convolutional framelet architecture with (a) $\Phi_1,\tilde\Phi_1$ and (b) $\Phi_2,\tilde\Phi_2$ in Eq.~\eqref{eq:Phi}. Here, each bar represents a channel so that  the number of channels is $s=3$ at the first layer decomposition. Similar encoder-decoder network are recursively applied to the three-channel feature map.}
	\label{fig:network_concept}
\end{figure}

Finally, a {\em properly designed} network refers to a network that satisfies the condition \eqref{eq:frame}, which is often called the frame condition \cite{ye2017deep}.  For example, the following two sets of pooling and unpooling operations satisfy the frame condition \eqref{eq:frame}:
\begin{eqnarray}\label{eq:Phi}
\Phi_1 = \tilde\Phi_1 = I_{n_1n_2} &,& \Phi_2 =\tilde\Phi_2= \begin{bmatrix} I_{n_1n_2}  & I_{n_1n_2} \end{bmatrix}
\end{eqnarray}
Here, the first set of basis ($\Phi_1,\tilde\Phi_1$) constitute a complete basis, whereas the second set of basis is redundant.  The corresponding multi-layer encoder-decoder network structures are illustrated in Figs.~\ref{fig:network_concept}(a) and (b), respectively.  Note that redundancies in the pooling/unpooling layers in $\Phi_2,\tilde\Phi_2$ are recognized as skipped connections.  In general, depending on the choice of pooling and unpooling matrices, various network forms can be derived.  See \cite{han2017framing} for more details. 

 In short, because of the redundancy in the Rx-Xmit (or Rx-SL) domain data, the associated Hankel matrix is low-ranked, which allows a direct interpolation of the Rx-Xmit (or Rx-SL) domain RF data using deep CNN.  Moreover, thanks to the nonlinearity and cascaded implementation, a properly designed neural network has exponentially increasing expressiveness suitable for interpolating diverse RF data.

\begin{figure*}[!htb]
	\centerline{\includegraphics[width=0.95\textwidth]{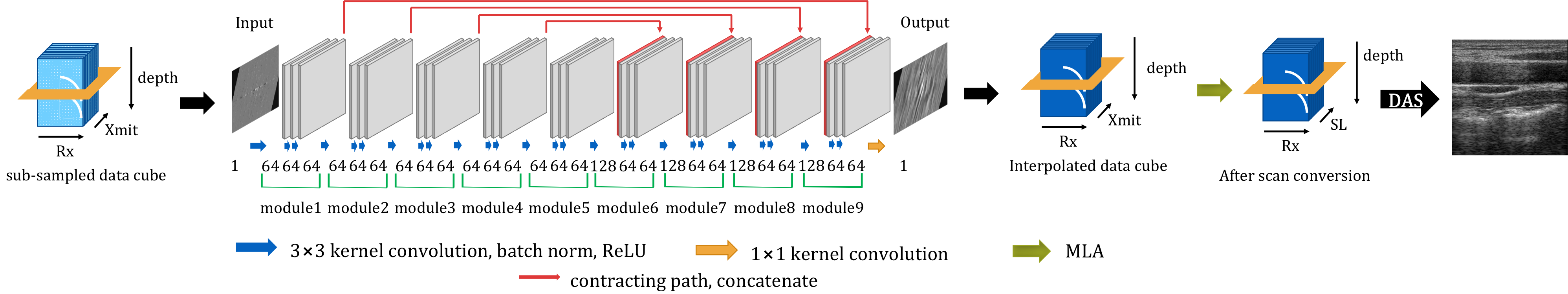}}
	\centerline{\mbox{(a)}}
	\vspace*{0.35cm}
		\centerline{\includegraphics[width=0.95\textwidth]{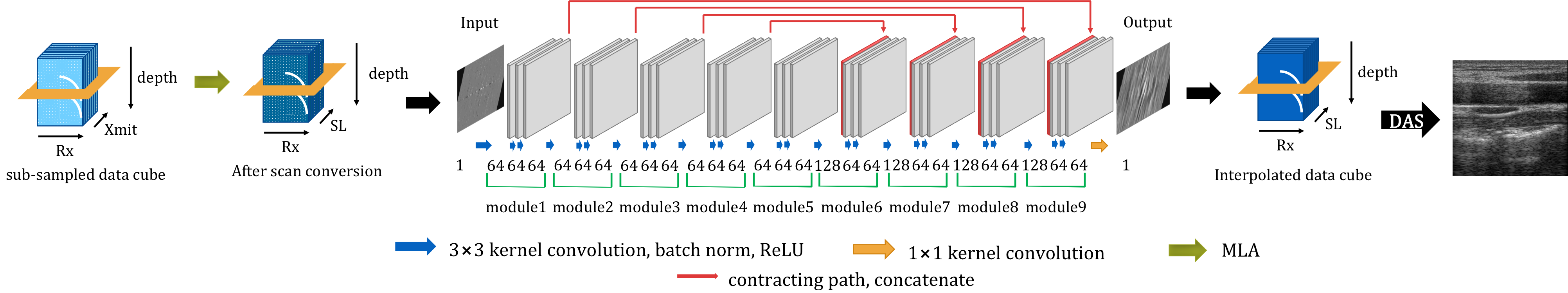}}
			\centerline{\mbox{(b)}}
	\caption{Our network architecture for RF-interpolation for (a) Rx sub-sampling, and (b) Rx-Xmit sub-sampling.}
	\label{fig:methodb}
\end{figure*}

\section{Method}
\label{sec:methods}

\subsection{Data set}

For experimental verification, multiple RF data were acquired with the E-CUBE 12R US system (Alpinion Co., Korea).  For data acquisition, we used a linear array transducer (L3-12H) with a center frequency of 8.48 MHz and a convex array transducer (SC1-4H) with a center frequency of 3.2 MHz.  The configuration of the probes is given in Table \ref{probe_config}.

\begin{table}[!hbt]
	\centering
	\caption{Probes Configuration}
	\label{probe_config}
	\resizebox{0.4\textwidth}{!}{
	\begin{tabular}{c|c|c}
\hline
		{Parameter} & {Linear Probe} & {Convex Probe} \\ \hline\hline
		Probe Model No.& L3-12H & SC1-4H \\
		Carrier wave frequency & 8.48 MHz & 3.2 MHz \\
		Sampling frequency & 40 MHz & 40 MHz \\
		No. of probe elements & 192 & 192 \\
		No. of Tx elements & 128 & 128 \\
		No. of Xmit events & 96 & 96 \\
		No. of Rx elements & 64 & 64 \\
		Elements pitch & 0.2 mm & 0.348 mm \\
		Elements width & 0.14 mm & 0.26 mm \\
		Elevating length & 4.5 mm & 13.5 mm \\ 
		Parallel beamforming & 4MLA & 4MLA \\ \hline
	\end{tabular}}
\end{table}

Using a linear probe, we acquired RF data from the carotid area from 10 volunteers.  The data consisted of 40 temporal frames per subject, providing 400 sets of Depth-Rx-Xmit data cube.  The dimension of each Rx-Xmit plane was $64\times96$.  A set of $30,000$ Rx-Xmit planes was randomly selected from the $400$ sets of data cubes, then divided into $25,000$ datasets for training and $5000$ datasets for validation.  The remaining dataset of $360$ frames was used as a test dataset.

In addition, we acquired $100$ frames of RF data from the abdominal regions of two subjects using a convex array transducer. The size of each Rx-Xmit plane in the convex probe case was also $64\times96$.  This dataset was only used for test purposes and no additional training of CNN was performed on it.  The convex dataset was used to verify the generalization power of the proposed algorithm.   

In both probes, the default setup for parallel beam formation was 4MLA, i.e, for each transmit event, four scan lines were synthesized.

\subsection{RF Sub-sampling Scheme}
\label{sec:RFsamScon}

For Rx sub-sampling experiments, the Rx data for each transmit event were randomly sub-sampled at a down-sampling factor of x$4$ or x$8$.  Since the receiver at the center gets the RF data from direct reflection as shown in Fig.~\ref{fig:theorya}(c), the RF data from the center of the active receivers set were always included to improve the performance.

We also considered sub-sampling in both the Xmit event and Rx direction. Specifically, the RF data was uniformly sub-sampled along the transmit event with a down-sampling factor of x2, which is followed by random subsampling along the Rx direction at a down-sampling ratio of 4 for each transmit event.  This scheme can potentially increase the number of temporal frames twice, and reduce the Rx power consumption by 4.

\subsection{Network Architecture}
For the Rx sub-sampling scheme, CNN was applied to $64\times96$ data in the Rx-Xmit plane. The interpolated data was later expanded to the $64\times 384$ Rx-SL plane using the parallel beam-forming scheme as shown in Fig.~\ref{fig:methodb}(a).  For Rx-Xmit sub-sampling, due to the uniform subsampling artifacts, the following scheme was found to be better.  Specifically, the RF data was first expanded into the $64\times 384$ Rx-SL plane using 8MLA.  CNN was then applied to the $64\times384$ Rx-SL plane as shown in Fig.~\ref{fig:methodb}(b).  

The proposed CNN is composed of convolution layers, batch normalization layers, ReLU layers and a contracting path with concatenation as shown in Figs.~\ref{fig:methodb}(a) and (b). Specifically, the network consists of 28 convolution layers composed of a batch normalization and ReLU except for the last convolution layer. The first 27 convolution layers use 3$\times$3 convolutional filters (i.e. the 2-D filter has a dimension of $3\times 3$), and the last convolution layer uses a 1$\times$1 filter. Four contracting paths with concatenation exist.  If the sub-sampling ratio is greater than 4, an additional convolutional layer with a 3$\times$3 filter, batch normalization layer, and ReLU layer is inserted in each module to enlarge the receptive field size. Note that the proposed architecture is a combination of two basic architectures in Figs.~\ref{fig:network_concept}(a) and (b).

The network was implemented with both TensorFlow \cite{abadi2016tensorflow} and MatConvNet \cite{vedaldi2015matconvnet} in the MATLAB 2015b environment to verify the platform-dependent sensitivity.  We found that with the same training strategy, the two implementations provided near identical results.  Specifically, for network training, the parameters were estimated by minimizing the $l_2$ norm loss function. The network was trained using a stochastic gradient descent with a regularization parameter of $10^{-4}$. The learning rate started from $10^{-7}$ and gradually decreased to $10^{-9}$. The weights were initialized using Gaussian random distribution with the Xavier method \cite{glorot2010understanding}. The number of epochs was $500$ for all down-sampling rates.  To avoid over-fitting during Rx-Xmit sub-sampling, the neural network was first trained with 4x Rx down-sampled data, and after $200$ epochs, the network was fine-tuned for expanded Rx-Xmit sub-sampled data.

The network was trained for random down sampling patterns to avoid bias to specific to sampling patterns. However, for each down-sampling scheme (e.g. x4, x8 and 4x2), a separate CNN was trained.

\begin{figure*}[!hbt]
	\centering
	\centerline{\includegraphics[width=0.9\textwidth]{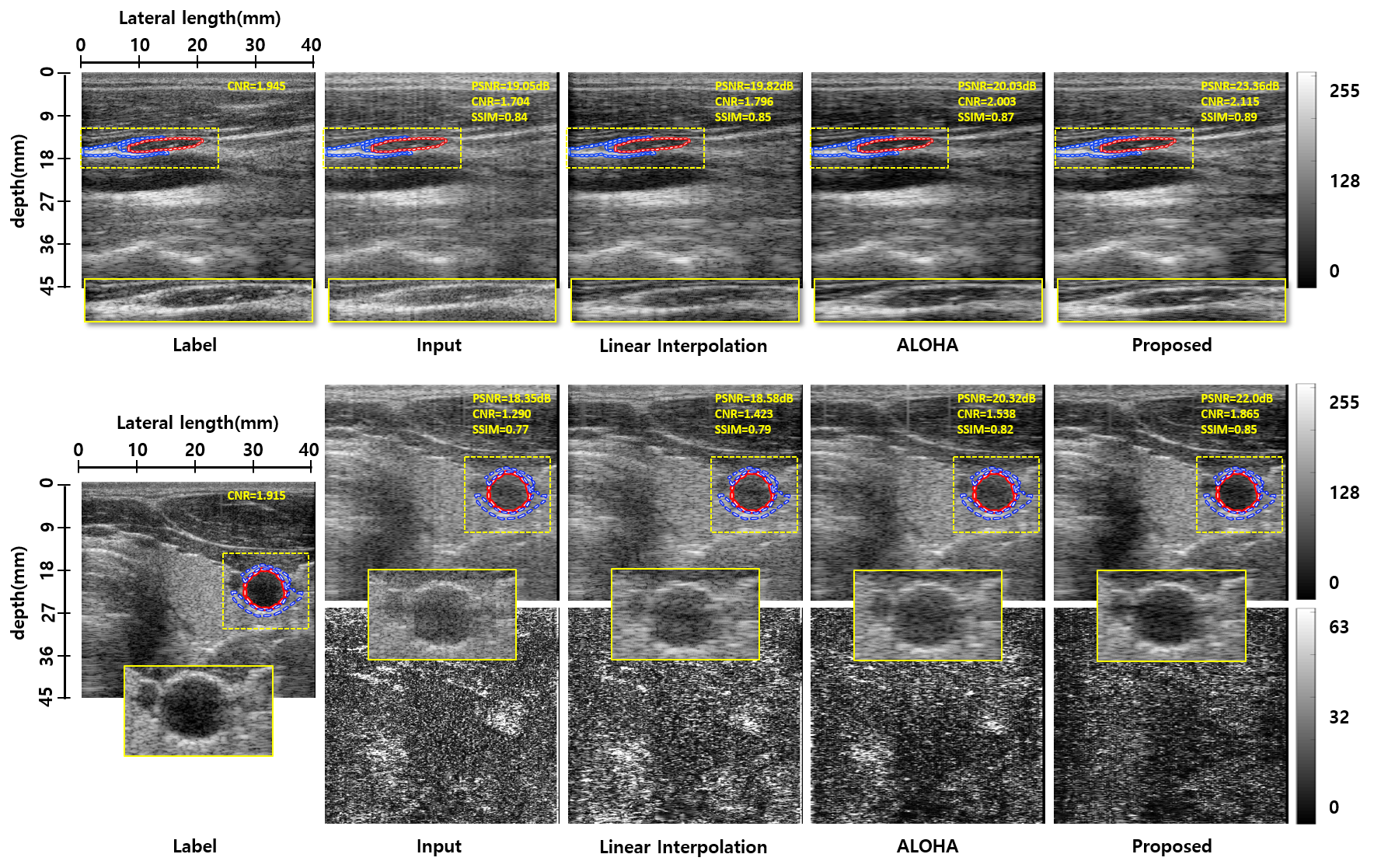}}
	\centerline{(a) x8 Rx sub-sampling results}\medskip
	\centerline{\includegraphics[width=0.9\textwidth]{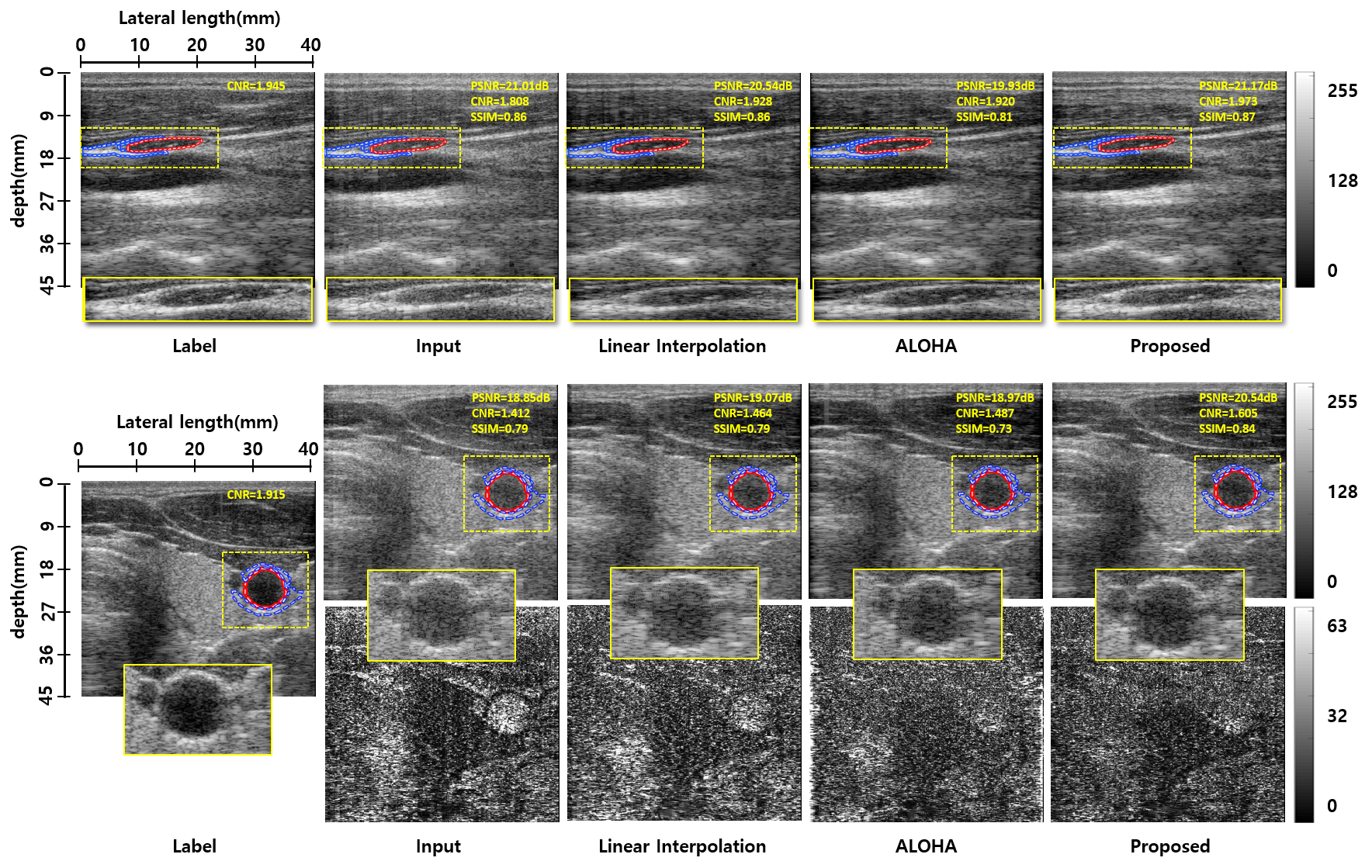}}
	\centerline{(b) $4\times2$ Rx-Xmit sub-sampling results}\medskip
	\caption{Reconstruction results of linear array transducer DAS beamformer B-mode images of carotid region from two sets of sub-sampled RF data.}
	\label{fig:resultsLin}
\end{figure*}

\subsection{Baseline algorithms}

For comparative studies, our CNN based interpolation was first compared with the linear interpolation results. Specifically, due to the irregular down-sampling pattern, the standard linear interpolation algorithm did not work, so we used the grid-based 3D interpolation function \textsf{griddata()} in MATLAB. However, at high sub-sampling ratios, the Rx-Xmit plane interpolation using \textsf{griddata()} still resulted in significant artifacts, so we used multiple temporal frames together for 3-D interpolation to improve the performance.

We also compared the annihilating filter-based low rank Hankel matrix approach (ALOHA) for RF interpolation \cite{jin2016compressive}, which also harnessed the correlation in the temporal direction in addition to the Rx-Xmit (or Rx-SL) domain redundancy.  Specifically, due to correlation in the temporal direction, the RF data from adjacent temporal frames showed some level of redundancy, which could be exploited by ALOHA \cite{jin2016general}.  More specifically, we could find the multi-frame annihilating filter relationship \cite{jin2016compressive}:
\begin{equation}
\label{eq:ALOHA6}
F_i \circledast K_j - F_i \circledast K_i = 0, \quad i\neq j ,
\end{equation}
where $F_i$ denotes the RF data in the Rx-Xmit data at the $i$-th frame, and $K_i$ are associated filters. Then, \eqref{eq:ALOHA6} can be represented in a matrix form:
\begin{equation}
\label{eq:ALOHA7}
\begin{bmatrix} \hank_{d_1,d_2}(F_i) & \hank_{d_1,d_2}(F_j) \end{bmatrix} \begin{bmatrix}
\overline\vec(K_i) \\
-\overline\vec(K_j)
\end{bmatrix}=0
\end{equation}
so that the extended Hankel matrix $\begin{bmatrix} \hank_{d_1,d_2}(F_i) & \hank_{d_1,d_2}(F_j) \end{bmatrix}$ is rank-deficient.  Similarly, we could construct an extended Hankel matrix using RF data from $N$-time frames:
\begin{equation}
\label{eq:ALOHA9}
\hank_{d_1,d_2|N}\left( \{F_i\}_{i=1}^N \right):=
\begin{bmatrix} \hank_{d_1,d_2}(F_1) & \cdots & \hank_{d_1,d_2}(F_N) \end{bmatrix} , \notag
\end{equation}
Due to the spatio-temporal correlation, the extended Hankel matrix still had a low rank \cite{jin2016compressive}.  Accordingly, the RF interpolation problem could be solved using the following low-rank Hankel
matrix completion problem \cite{jin2016compressive}:
\begin{eqnarray}
\label{eq:ALOHA10}
\min_{\{M_i\}_{i=1}^N } && \|\hank_{d_1,d_2|N}\left( \{M_i\}_{i=1}^N \right)\|_*\\
\mbox{subject to} && P_{\Lambda}[F_i] = P_\Lambda[M_i], \quad i=1,\cdots, N \notag,
\end{eqnarray}
where $\Lambda$ denotes the indices of the measured RF data and $P_{\Lambda}$ denotes the projection to the index $\Lambda$.  The optimization problem \eqref{eq:ALOHA10} can be solved using an alternating direction method of multiplier (ADMM) after matrix factorization \cite{jin2016compressive}.

The computational complexity of ALOHA is mainly determined by the number of annihilating filter sizes and the temporal frames.  In our case, we used 10 temporal frames and an annihilating filter size of $7\times 7$.  The rigorous complexity analysis of ALOHA is provided in Appendix A.

\subsection{Performance Metrics}

 To quantitatively show the advantages of the proposed deep learning method, we used the contrast-to-noise ratio (CNR) \cite{BiomedicalImageAnalysis}, peak-signal-to-noise ratio (PSNR), structure similarity (SSIM) \cite{1284395} and the reconstruction time.  
 
 The CNR is measured for the background ($B$) and anechoic structure ($aS$) in the image, and is quantified as
 \begin{equation}
 {\hbox{CNR}}(B,aS) = \frac{|\mu_{B}-\mu_{aS}|}{\sqrt{\sigma^2_{B} + \sigma^2_{aS}}},
 \end{equation}
 where $\mu_{B}$, $\mu_{aS}$, and $\sigma_{B}$, $\sigma_{aS}$ are the local means, and the standard deviations of the background ($B$) and anechoic structure ($aS$) \cite{BiomedicalImageAnalysis}. 
 
 The PSNR and SSIM index are calculated on label ($F$) and reconstruction ($\tilde F$) images of common size  $n_1\times n_2$ as

\begin{equation}
{\hbox{PSNR}}(F,\tilde F) = 10 \ensuremath{\log_{10}} \left(\frac{ n_1n_2 R_{\max}^2}{\|F-\tilde F\|_F^2}\right),
\end{equation}
where $\|\cdot\|_F$  denotes the Frobenius norm and $R_{\max}=2^{(\#bits\ per\ pixel)}-1$ is the dynamic range of pixel values (in our experiments this is equal to $255$),
and
\begin{equation}
{\hbox{SSIM}}(F,\tilde F) = \frac{(2\mu_{F}\mu_{\tilde F}+c_{1})(2\sigma_{F,\tilde F} +c_{2})}{(\mu^{2}_{F}+\mu^{2}_{\tilde F}+c_{1})(\sigma^{2}_{F}+\sigma^{2}_{\tilde F} +c_{2})},
\end{equation}
where  $\mu_{F}$, $\mu_{\tilde F}$, $\sigma_{F}$, $\sigma_{\tilde F}$, and $\sigma_{F,\tilde F}$ are the local means, standard deviations, and cross-covariance for images $F$ and $\tilde F$ calculated for a radius of $50$ units.  The default values of $c_{1}=(k_{1}R_{max})^{2}$, $c_{2}=(k_{2}R_{max})^{2}$, $k_{1}=0.01$ and $k_{1}=0.03$.

  \begin{figure}[!htb]
 	\centerline{\includegraphics[width=8.5cm, height=2cm]{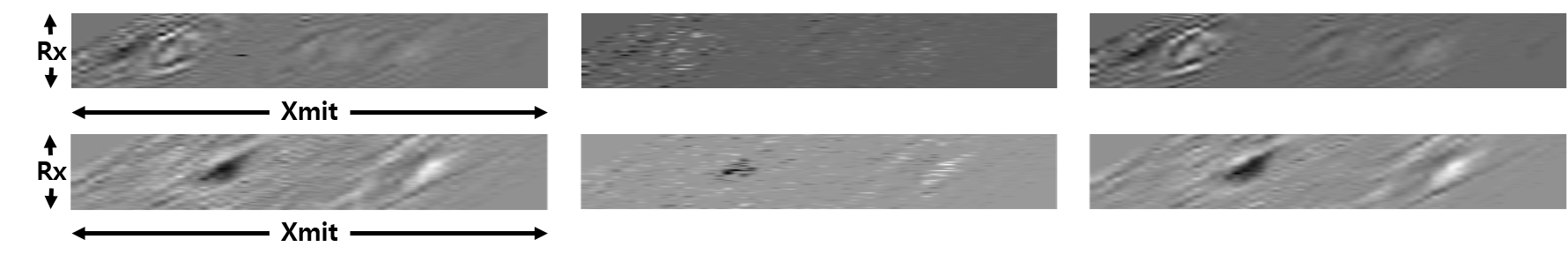}}
	\vspace*{-0.45cm}
	\centerline{\mbox{(a)}}
	\vspace*{0.25cm}

 	\centerline{\includegraphics[width=8.5cm, height=2cm]{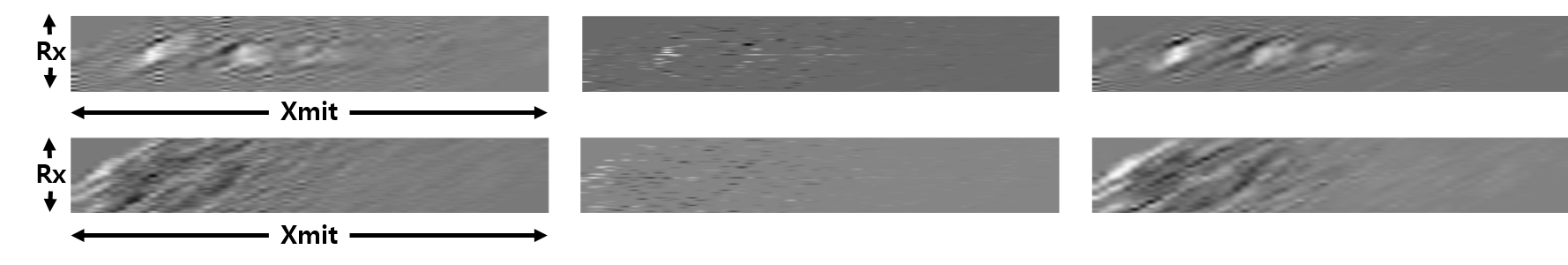}}
	\vspace*{-0.45cm}
	\centerline{\mbox{(b)}}
	\vspace*{0.25cm}
 	\centerline{\includegraphics[width=8.5cm, height=2cm]{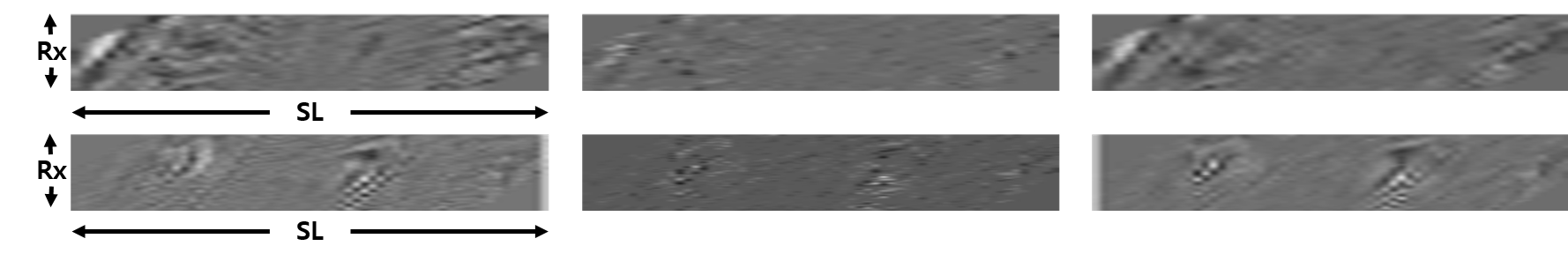}}
	\vspace*{-0.45cm}
	\centerline{\mbox{(c)}}
 	\caption{Rx-SL coordinate RF data for (a) x4 Rx down-sampling, (b) x8 Rx down-sampling, (c) 2x4 Rx-Xmit down-sampling. Label denotes the fully sampled RF data, and output refers to the interpolated output from input using the proposed method.}
 	\label{fig:results_RxSC}
 \end{figure}

 \begin{figure*}[!hbt]
	\centerline{\includegraphics[width=7cm]{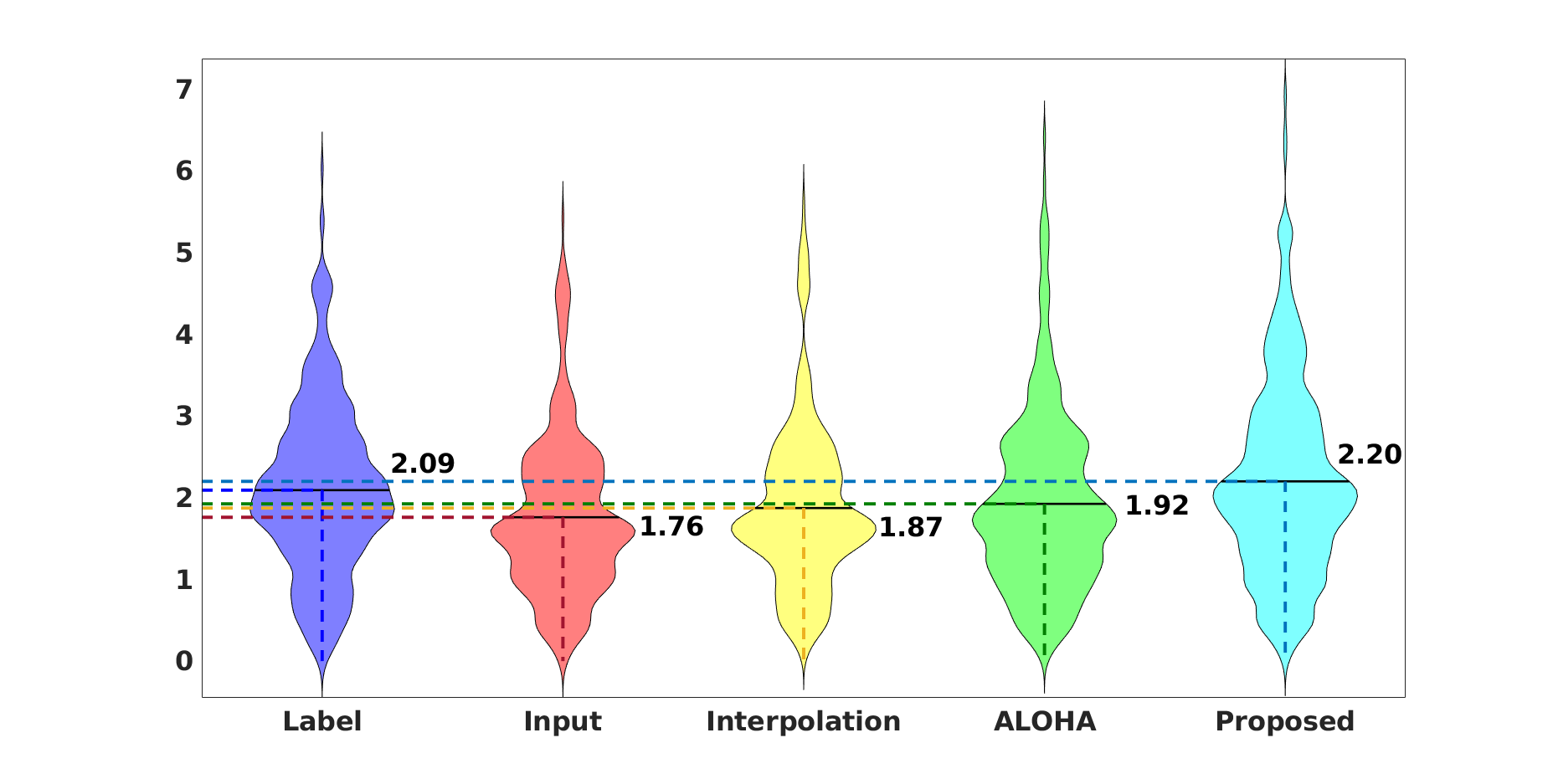}\hspace*{-0.5cm}\includegraphics[width=7cm]{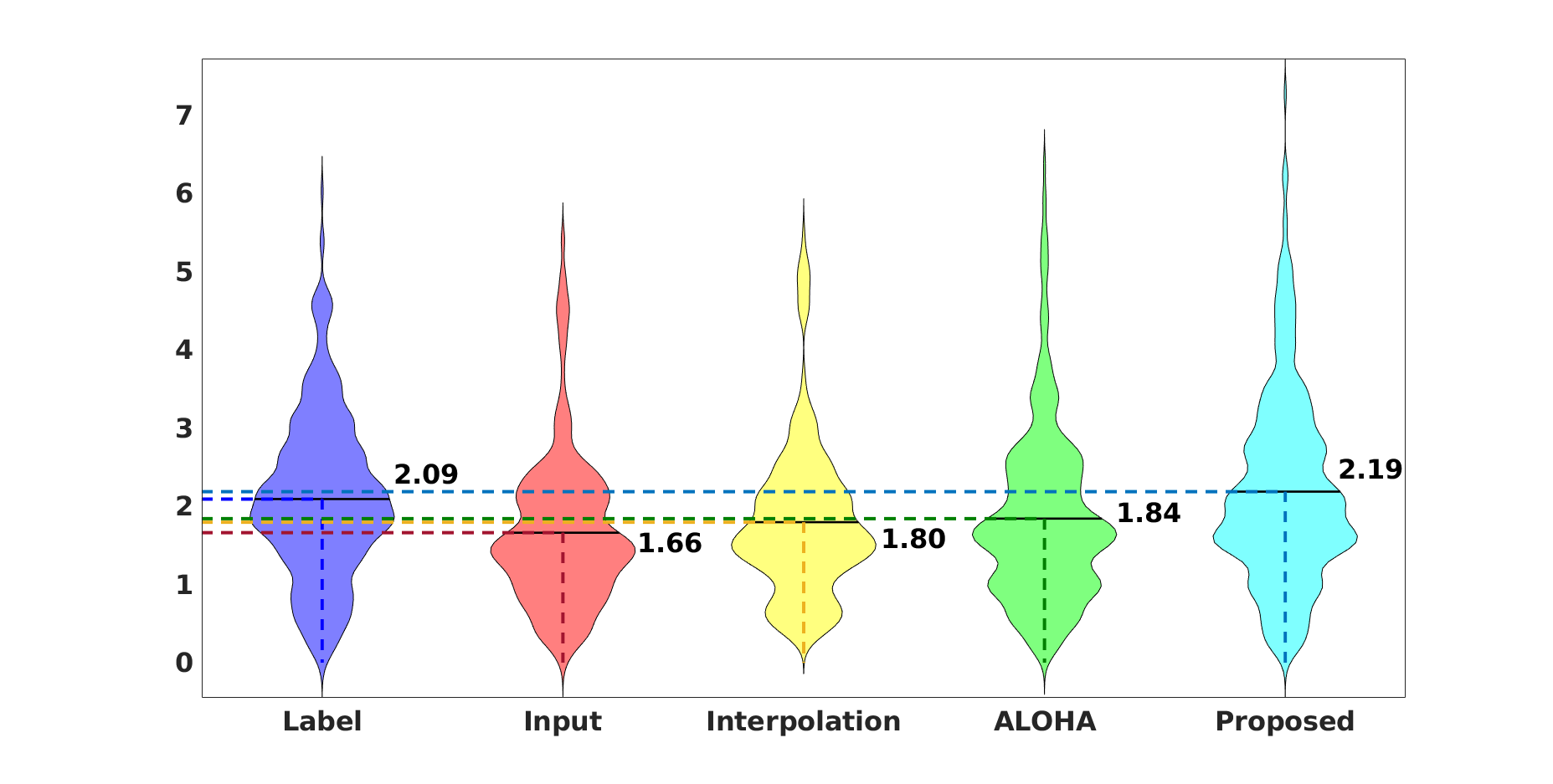}\hspace*{-0.5cm}\includegraphics[width=7cm]{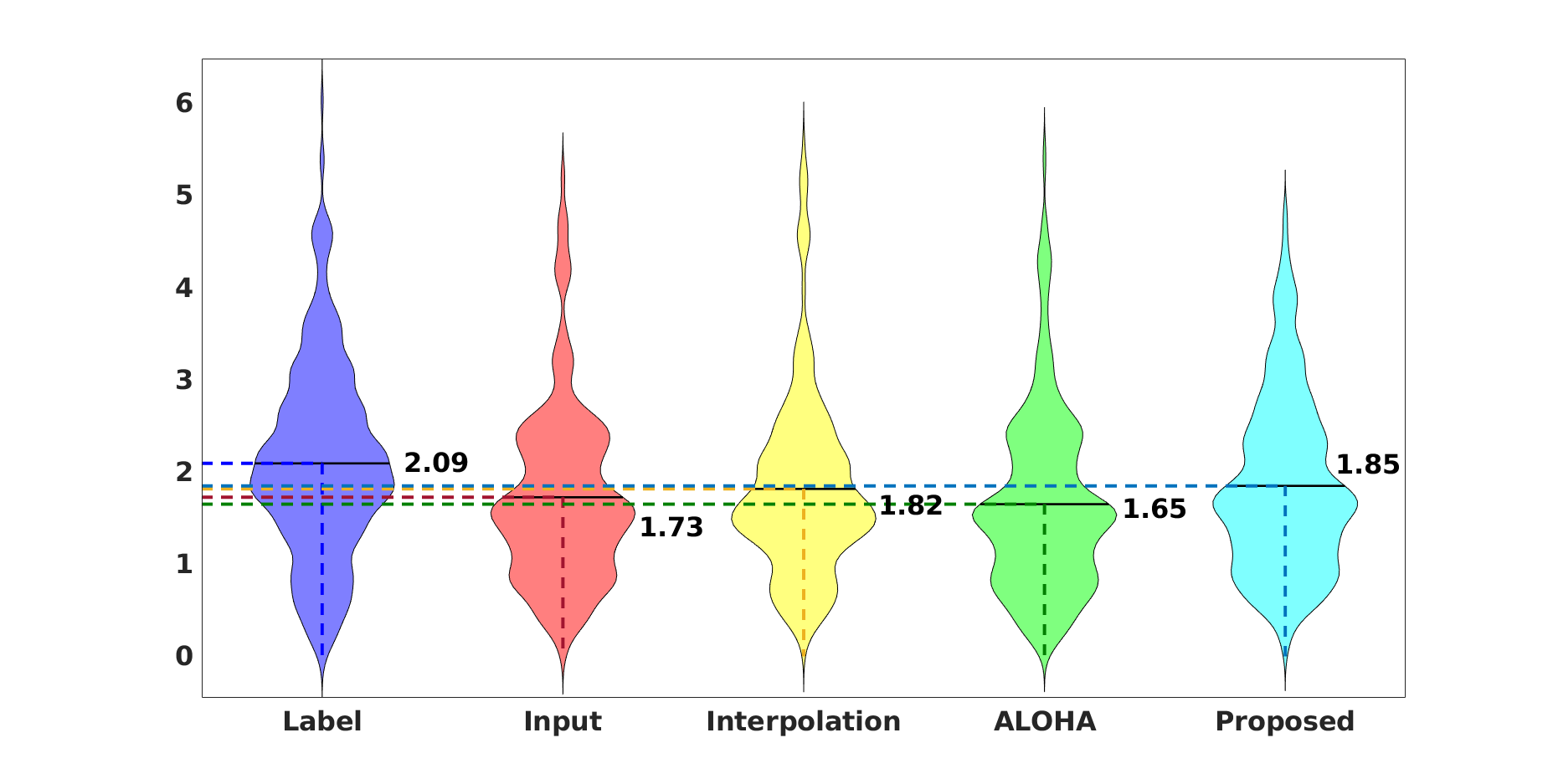}}
	\centerline{\mbox{(a) CNR value distribution }}
	\vspace*{0.3cm}
	\centerline{\includegraphics[width=7cm]{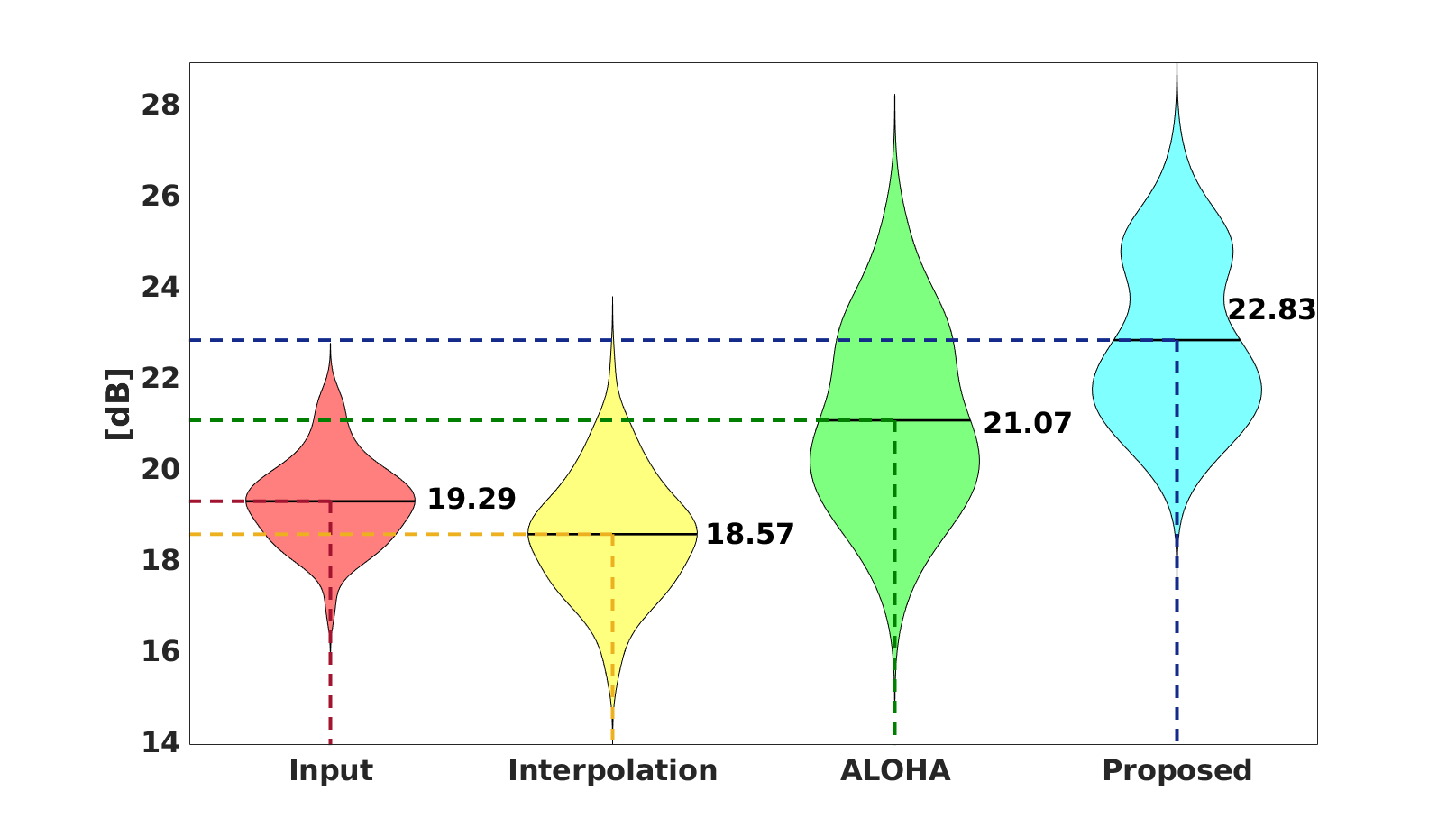}\hspace*{-0.5cm}\includegraphics[width=7.cm]{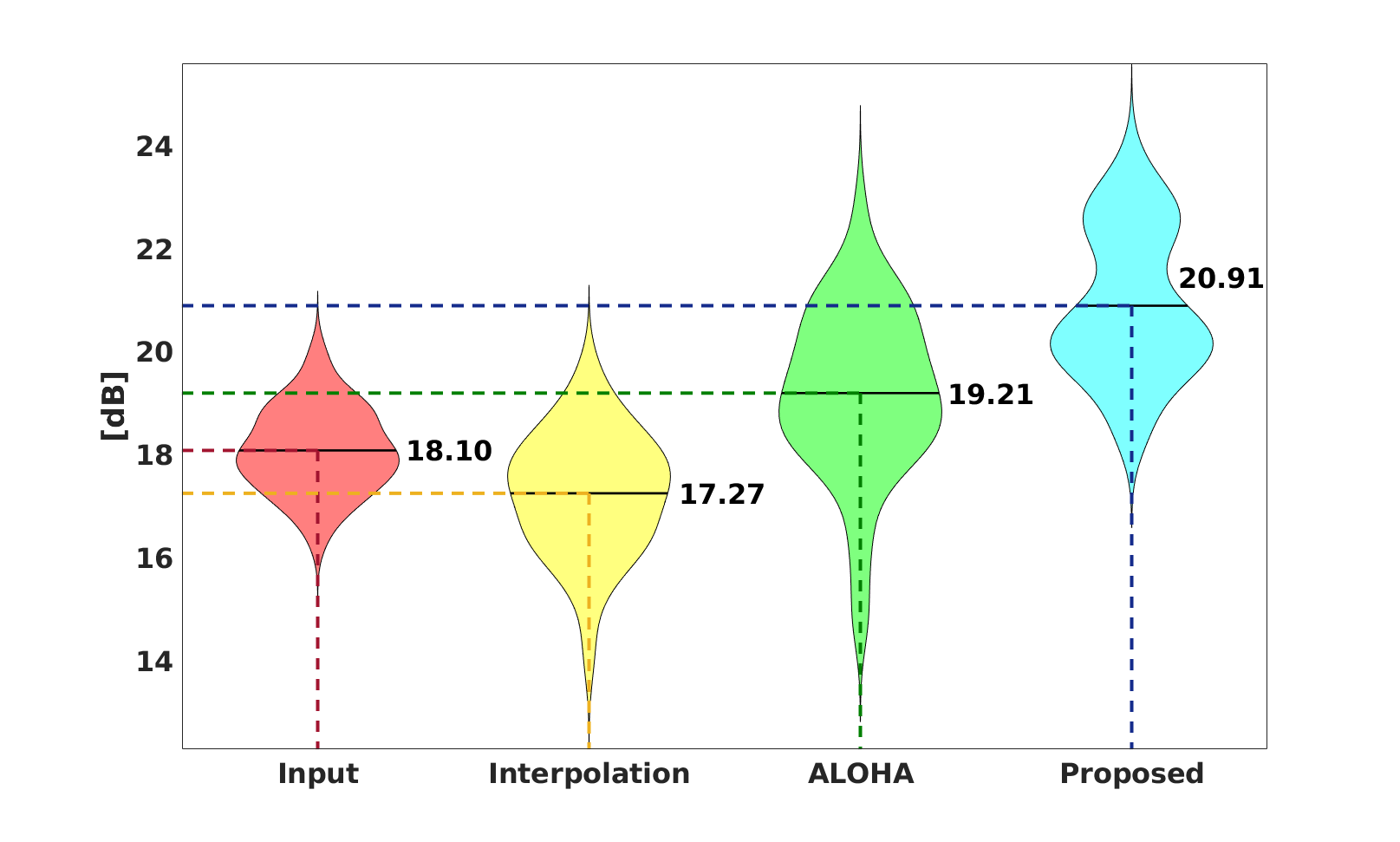}\hspace*{-0.5cm}\includegraphics[width=7.cm]{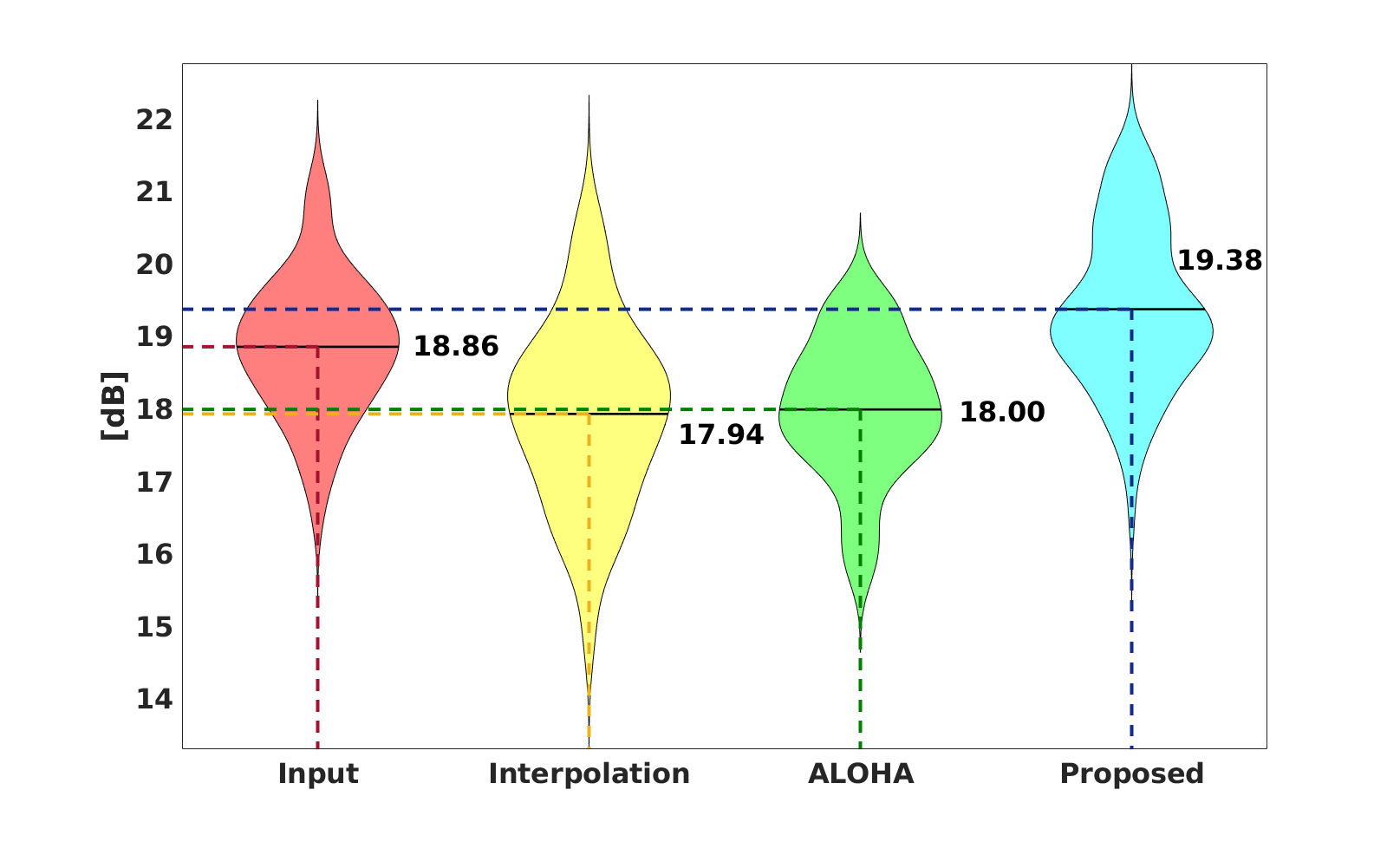}}
	\centerline{\mbox{(b) PSNR value distribution (dB)}}
	\vspace*{0.3cm}
	\centerline{\includegraphics[width=7cm]{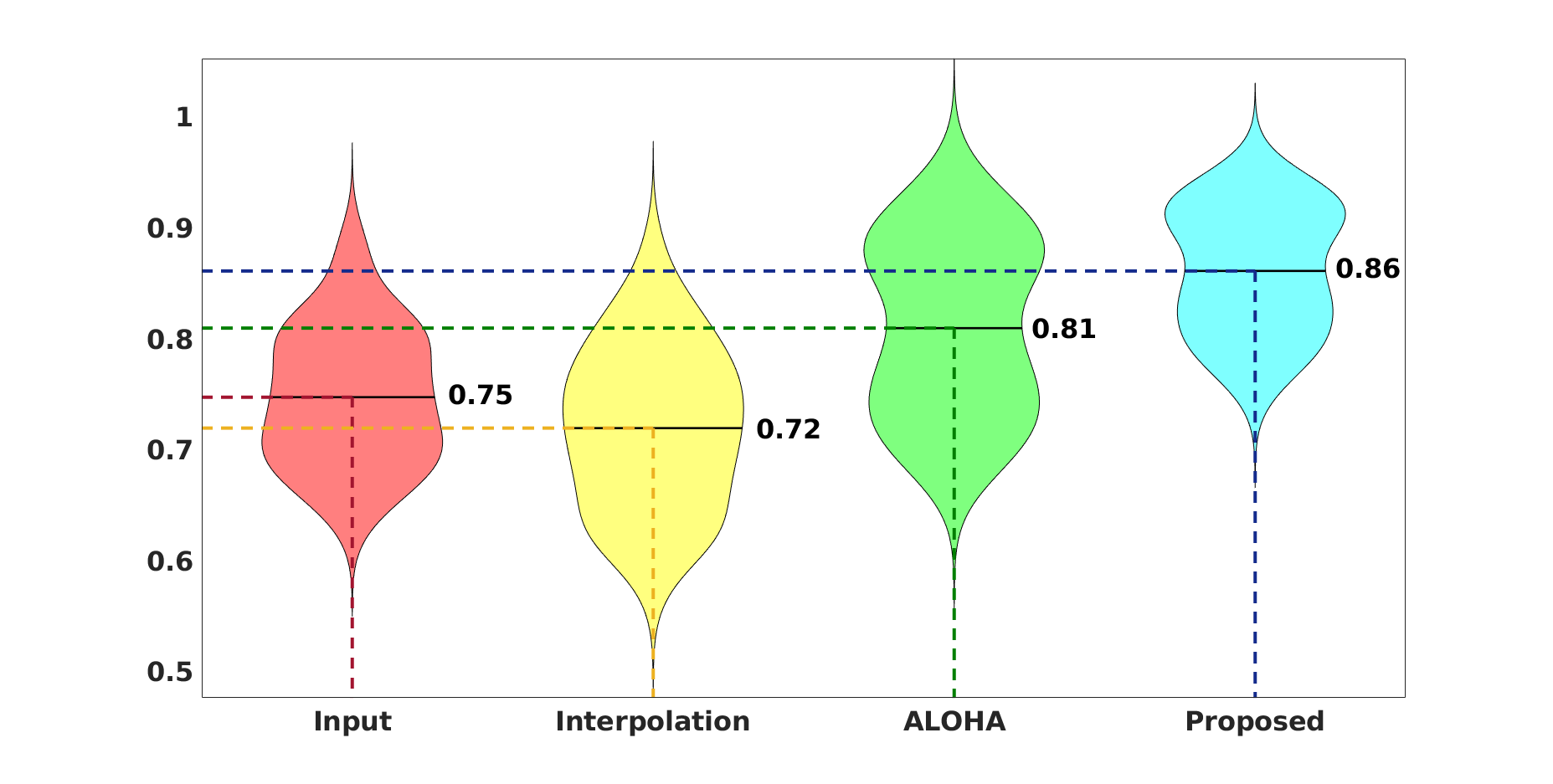}\hspace*{-0.5cm}\includegraphics[width=7.cm]{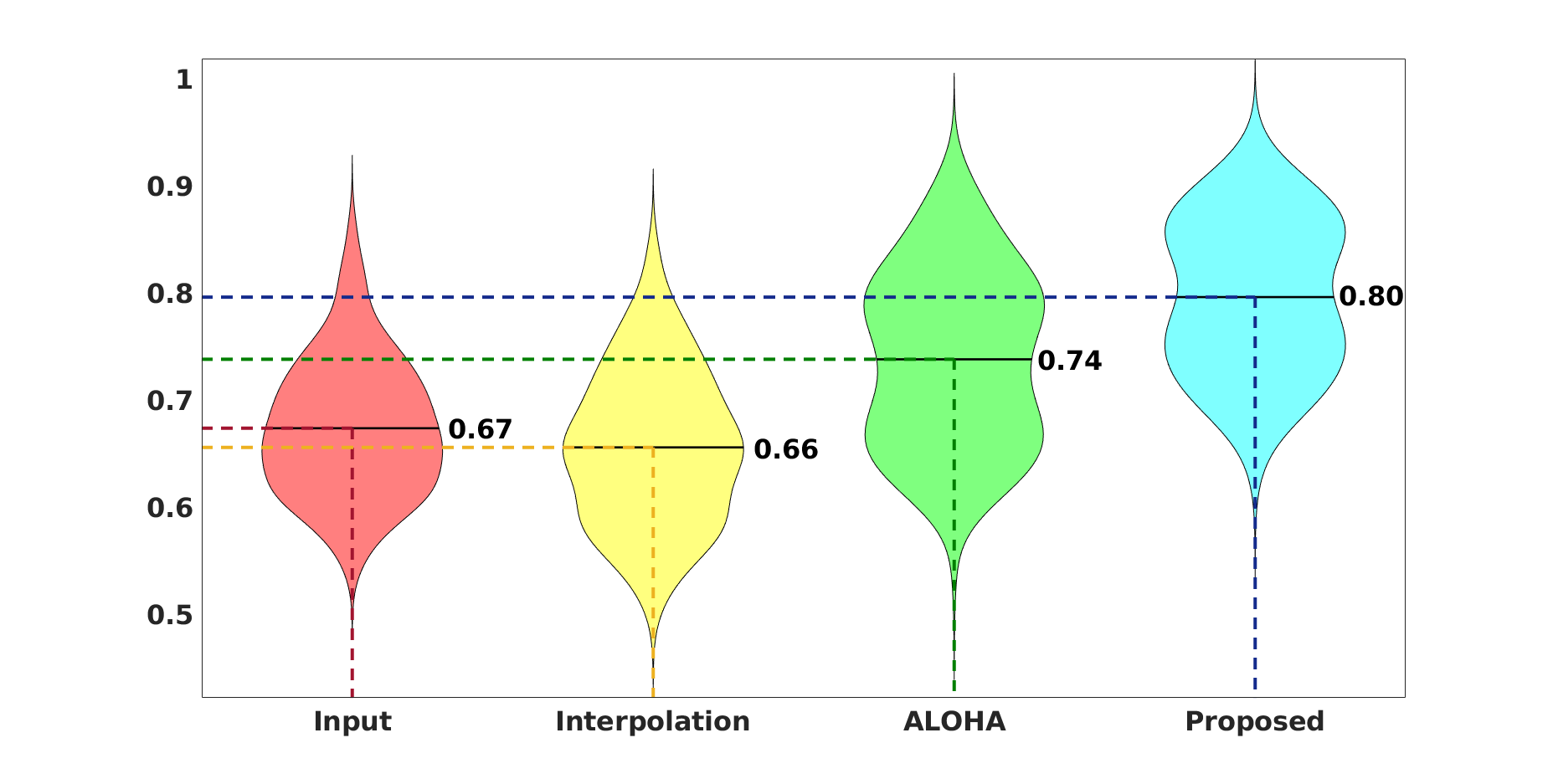}\hspace*{-0.5cm}\includegraphics[width=7.cm]{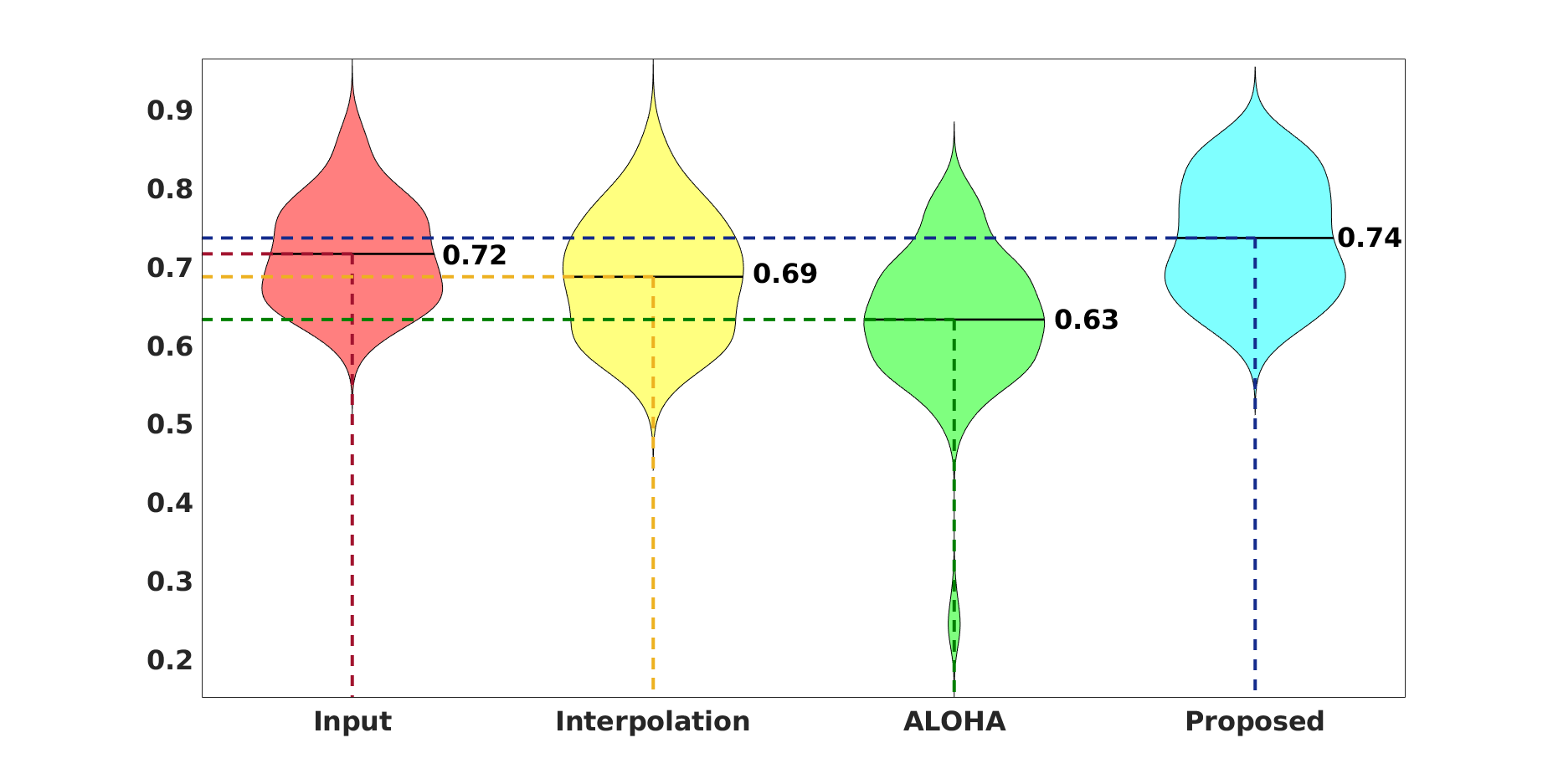}}
		\centerline{\mbox{(c) SSIM value distribution }}
	\caption{Carotid region linear probe B-mode Reconstruction CNR, SSIM and PSNR value distribution of 360 images from various RF sub-sampling scheme: (First column) x4 Rx sub-sampling, (middle column) x8 Rx sub-sampling, (last column) 4x2 Rx-Xmit sub-sampling.}
	\label{fig:results_CNR_PSNR_SSIM_Lin}
\end{figure*}

\section{Experimental Results}
\label{sec:results}

 \subsection{Linear Array Transducer Experiments}
 
  Fig.~\ref{fig:resultsLin} show the DAS beam-former output images for $8\times$ Rx and $4\times2$ Rx-Xmit down-sampling schemes.  Our method significantly improves the visual quality of the input image and outperforms other methods by eliminating line and block artifacts.  From difference images in Fig.~\ref{fig:resultsLin}, it is evident that under both down-sampling schemes, the proposed method approximate both the near and the far field Rx-Xmit planes with equal efficacy and only few structural details are discernible. On the other hand, linear interpolation exhibits horizontal blur artifacts especially at the far field region as shown in the difference images.

  Fig.~\ref{fig:results_RxSC} illustrates representative examples of Rx-SL coordinate data from the linear transducer for three different RF down-sampling schemes.  The proposed CNN-based interpolation successfully reconstructs the missing RF data in all down-sampling schemes.  CNN automatically identifies the missing RF data and approximates it with available neighboring information.  It is noteworthy that in the down-sampling schemes, the proposed method can efficaciously interpolates the missing data from as little as only $12.5$\% RF-data.

  We compared the CNR, PSNR, and SSIM distributions of reconstructed B-mode images obtained from 360 test frames of a linear array.  In Fig.~\ref{fig:results_CNR_PSNR_SSIM_Lin}(a), compared to the linear interpolation and ALOHA, the proposed method achieved average CNR values of $2.20$, $2.19$, and $1.85$ in $\times4$ Rx, $\times8$ Rx and $4\times2$ Rx-Xmit sampling schemes, respectively, These values are $25.00\%$, $31.93\%$ and $6.94\%$ higher than the input, $17.65\%$, $21.67\%$ and $1.65\%$ higher than the linear interpolation results, and $14.58\%$, $19.02\%$ and $12.12\%$ higher than the ALOHA results.

Fig.~\ref{fig:results_CNR_PSNR_SSIM_Lin}(b) compares the PSNR distribution in $4\times$ Rx, $8\times$ Rx, and $4\times 2$ Rx-Xmit sub-sampling schemes. Compared to linear interpolation, the proposed deep learning method showed $4.26$dB, $3.64$dB, and $1.44$dB improvement on average for $4\times$ Rx, $8\times$ Rx, and $4\times 2$ Rx-Xmit sub-sampling schemes, respectively. In comparison to ALOHA, the proposed deep learning method showed $1.76$dB, $1.7$dB, and $1.38$dB improvement on average for $4\times$ Rx, $8\times$ Rx, and $4\times 2$ Rx-Xmit sub-sampling schemes, respectively.

Unlike linear interpolation and ALOHA, which rely on temporal correlation of the RF data in multiple frames, the proposed method reconstructs each frame individually.  Therefore, the structure similarity (SSIM) of the DAS beam-former images in the proposed algorithm is significantly high.  Fig.~\ref{fig:results_CNR_PSNR_SSIM_Lin}(c) compare the SSIM in $4\times$ Rx, $8\times$ Rx, and $4\times 2$ Rx-Xmit sub-sampling schemes. Compared to linear interpolation, the proposed deep learning method showed $19.44\%$, $21.21\%$, and $7.25\%$ improvement in $4\times$ Rx, $8\times$ Rx, and $4\times 2$ Rx-Xmit sub-sampling schemes, respectively.  In comparison to ALOHA, the proposed deep learning method showed $6.17\%$, $8.11\%$, and $17.46\%$ improvement in $4\times$ Rx, $8\times$ Rx, and $4\times 2$ Rx-Xmit sub-sampling schemes, respectively.

Another important advantage of the proposed method is the run-time complexity. Although training required 96 hours for 500 epochs using Tensorflow, once training was completed, the reconstruction time for the proposed deep learning method was several orders of magnitude faster than those for ALOHA and linear interpolation (see Table \ref{recon_time}).

\begin{table}[!hbt]
	\centering
	\caption{Average reconstruction time (milliseconds) for each Rx-Xmit planes}
	\label{recon_time}
	 \resizebox{0.45\textwidth}{!}{
	\begin{tabular}{c|c|c|c}
		\hline
		Sub-sampling scheme & griddata() & ALOHA & Proposed  \\ \hline\hline
 	x4 Rx & 41.0 & 65.5 & 1.0 \\
x8 Rx & 31.1 & 58.9 & 9.8 \\ 
4x2 Rx-Xmit & 38.2 & 107.0 &  1.0 \\ \hline
	\end{tabular}
	}
\end{table}

 \subsection{Convex Array Transducer Experiments}
 
 \begin{figure*}[htb]
 	\centering
 	\centerline{\includegraphics[width=0.45\textwidth]{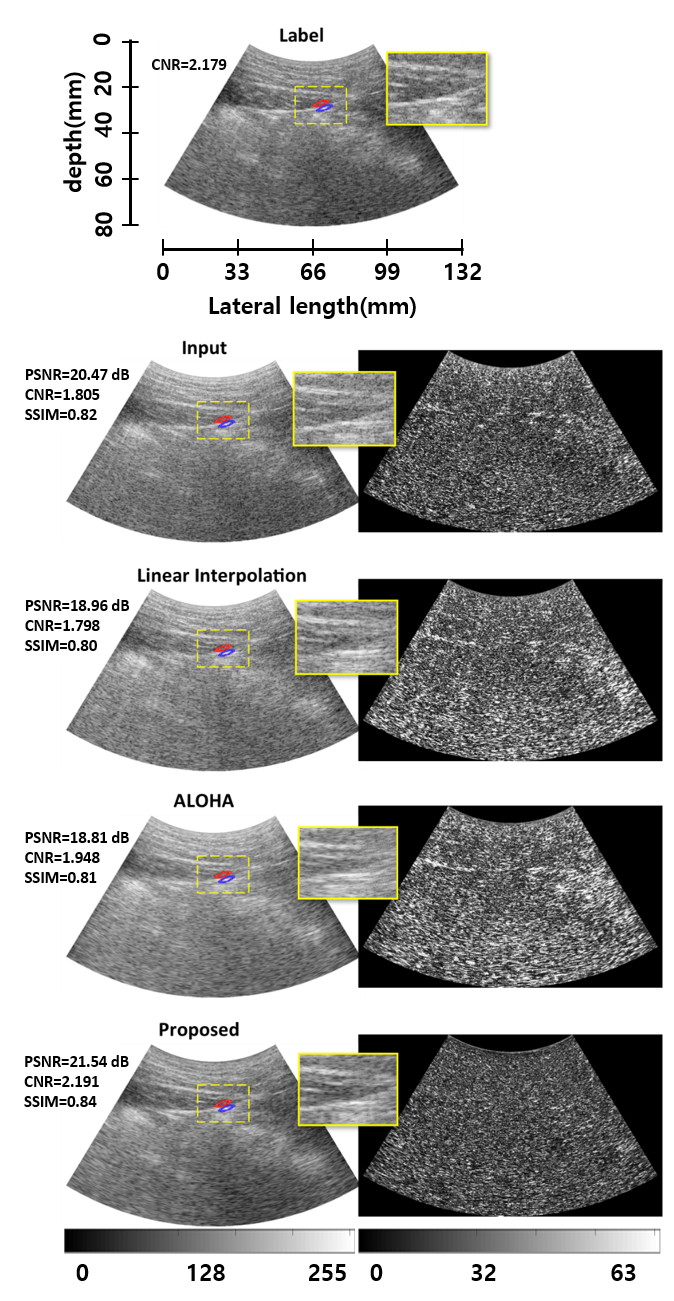} \hspace{1cm}\includegraphics[width=0.45\textwidth]{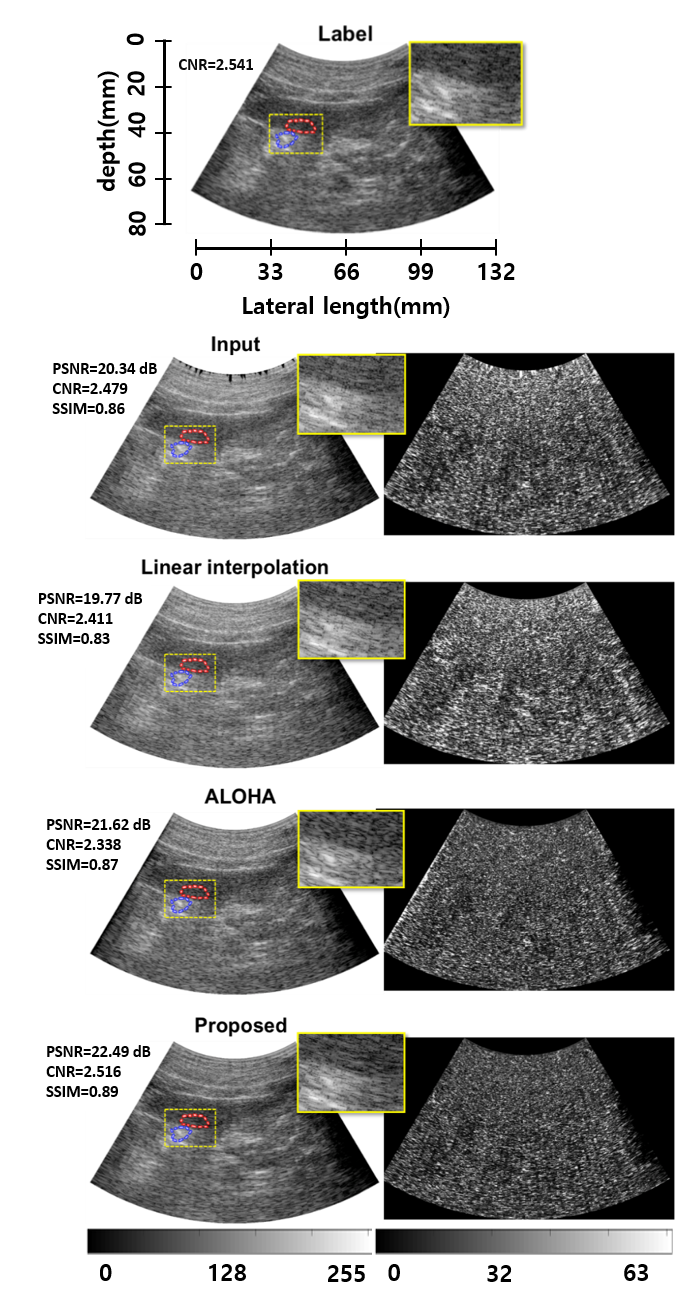}}
 	\centerline{(a) x8 Rx subsampling results \hspace{4cm} (b) $4\times2$ Rx-Xmit sub-sampling results}
 	\caption{Reconstruction results for abdominal region (liver) from sub-sampled RF data using the convex probe.}
 	\label{fig:resultsCon}
 \end{figure*}

 The trained CNN from the linear array transducer was applied to the circular array transducer.   Fig.~\ref{fig:resultsCon} shows the RF interpolated DAS beam-former output images for $8\times$ Rx and $4\times2$ Rx-Xmit down-sampling schemes.  Excellent image quality was obtained with the proposed method, which outperformed the existing algorithms. Moreover, many artifacts shown in the zoomed part of input images were successfully removed.  Although the network was trained using linear array transducers, the performance improvement using the proposed algorithm in convex array transducer was similar to that for the linear array experiment.

\begin{figure*}[htb]
	\centerline{\includegraphics[width=7cm]{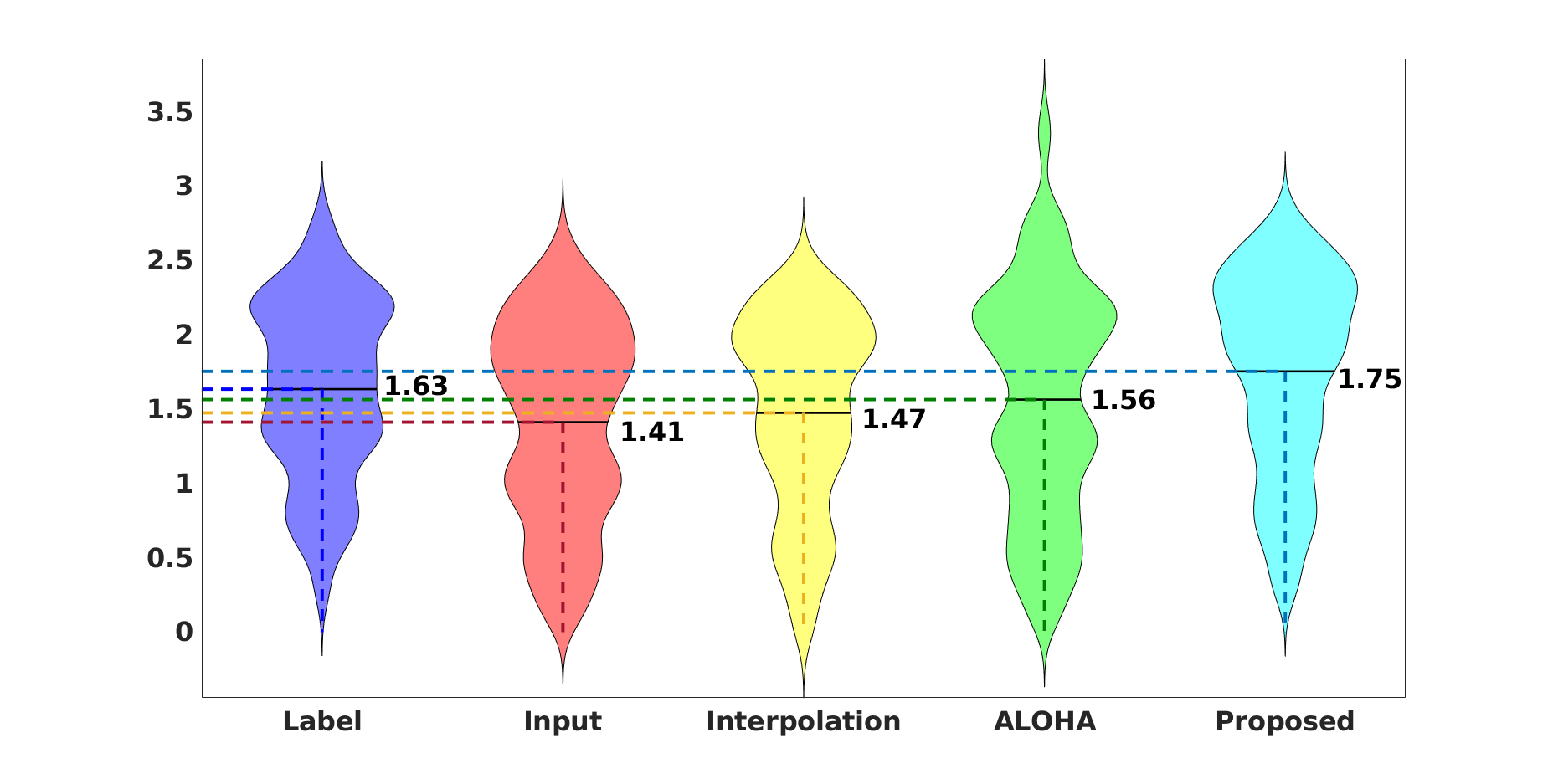}\hspace*{-0.5cm}\includegraphics[width=7cm]{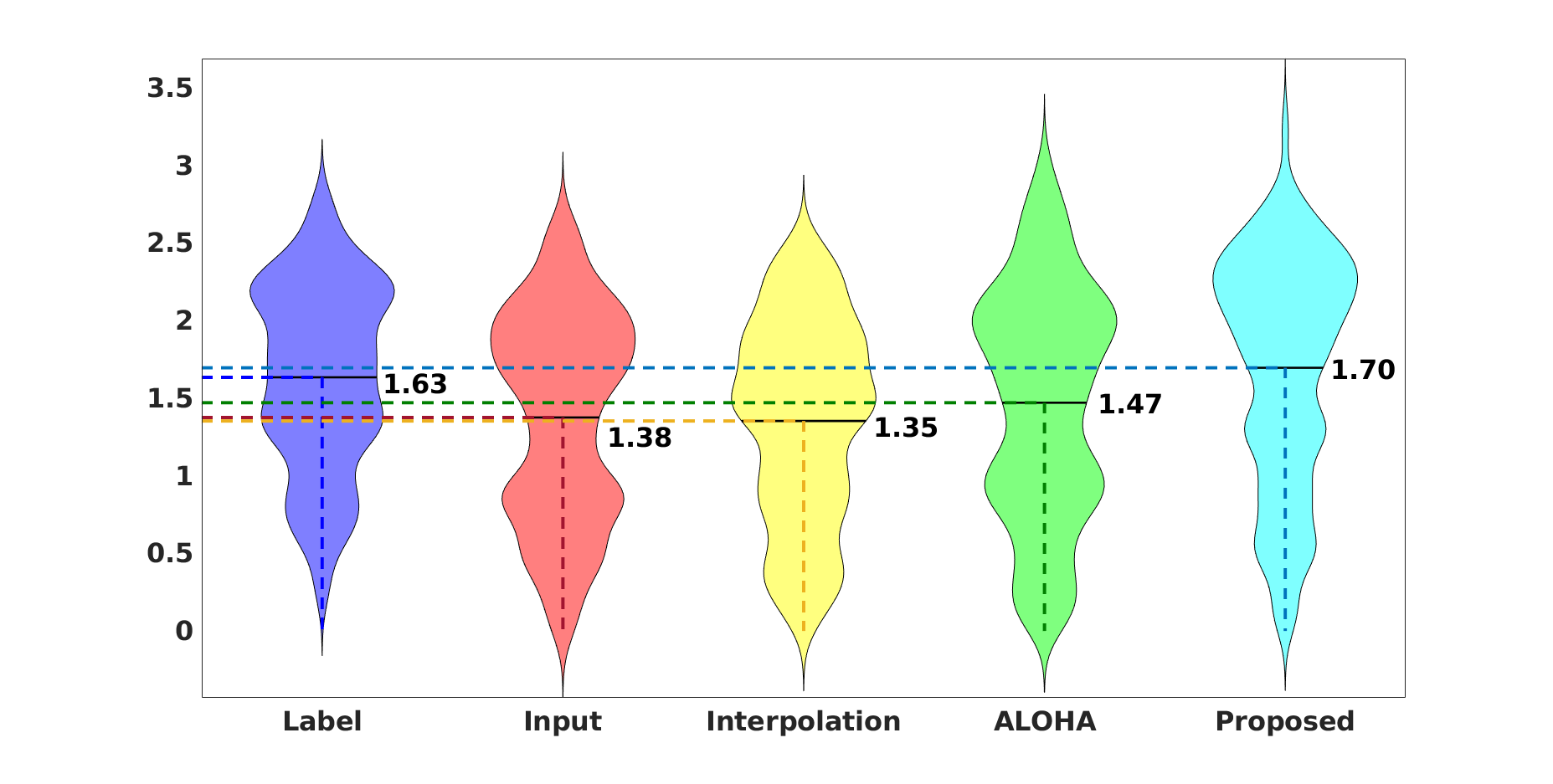}\hspace*{-0.5cm}\includegraphics[width=7cm]{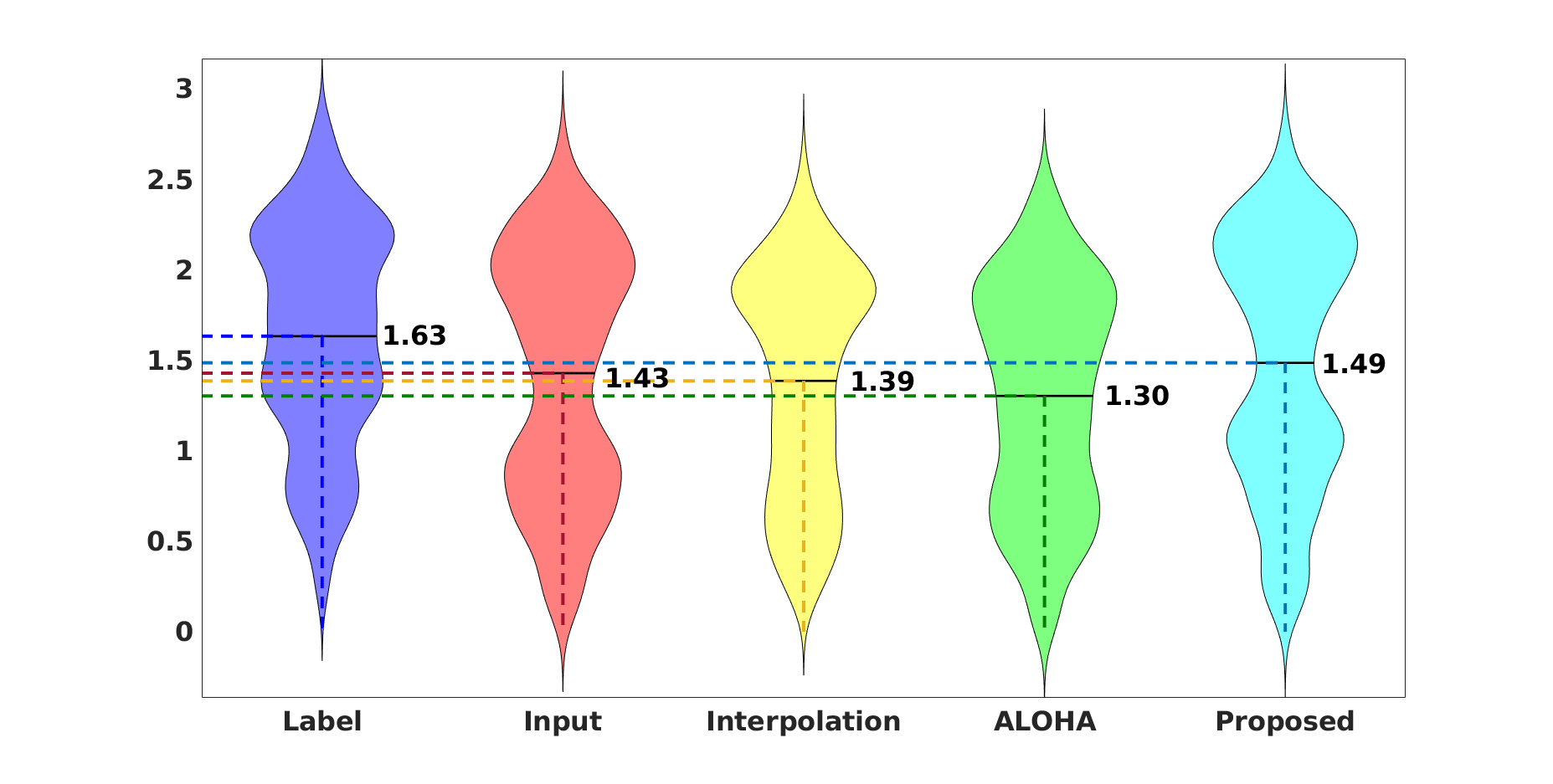}}
	\centerline{\mbox{(a) CNR value distribution (dB)}}
	\vspace*{0.3cm}
	\centerline{\includegraphics[width=7cm]{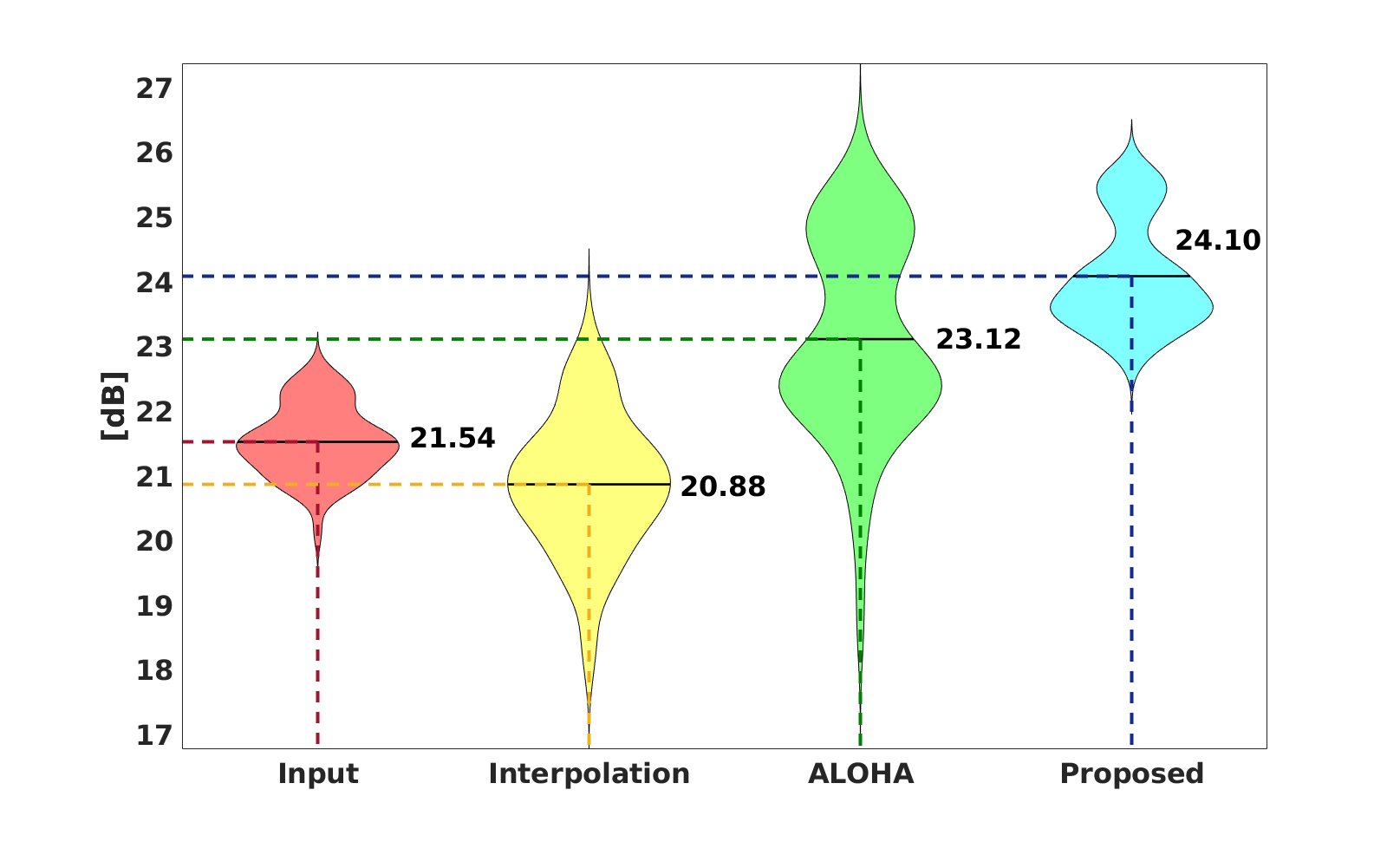}\hspace*{-0.5cm}\includegraphics[width=7.cm]{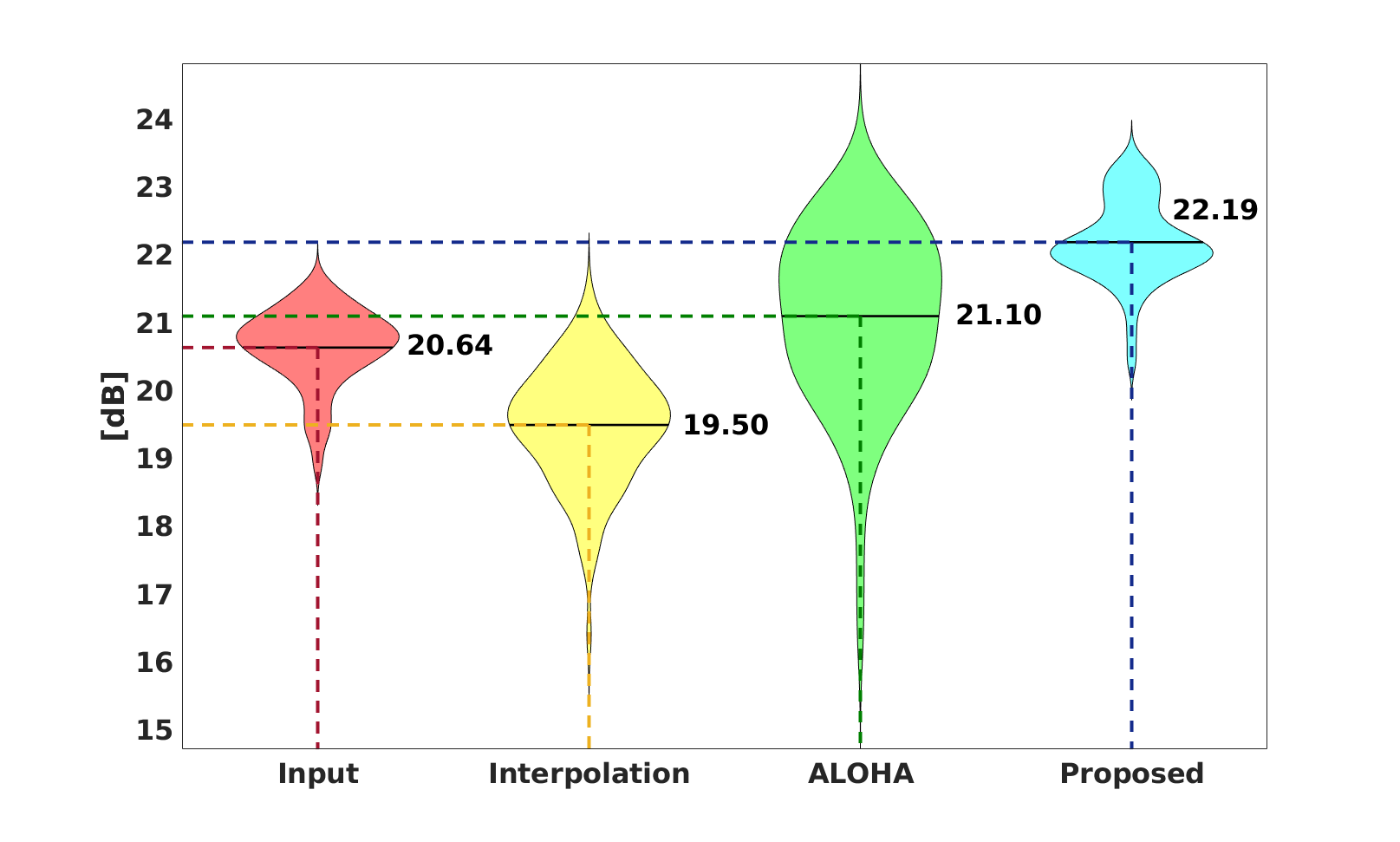}\hspace*{-0.5cm}\includegraphics[width=7.cm]{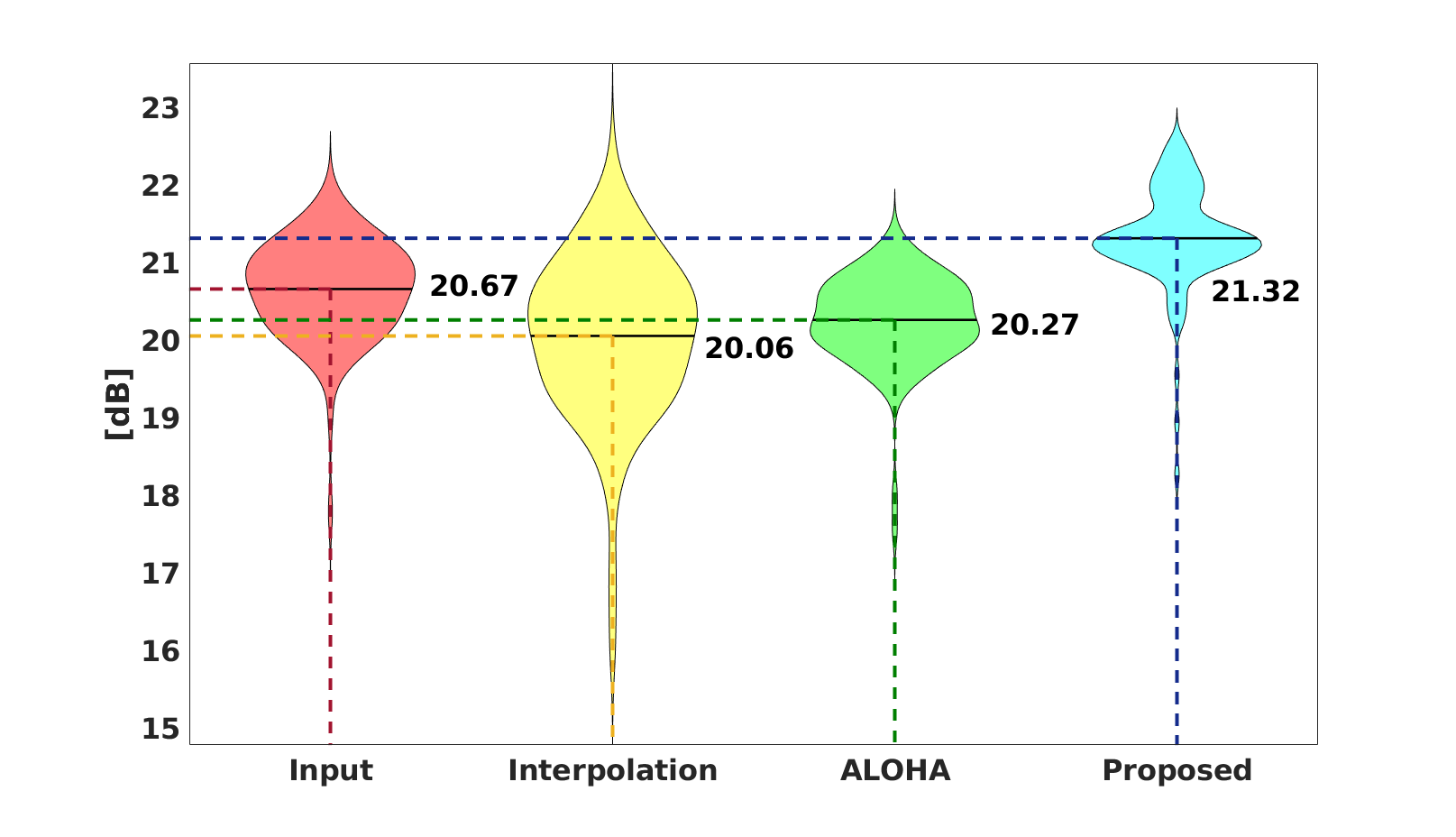}}
	\centerline{\mbox{(b) PSNR value distribution (dB)}}
	\vspace*{0.3cm}
	\centerline{\includegraphics[width=7cm]{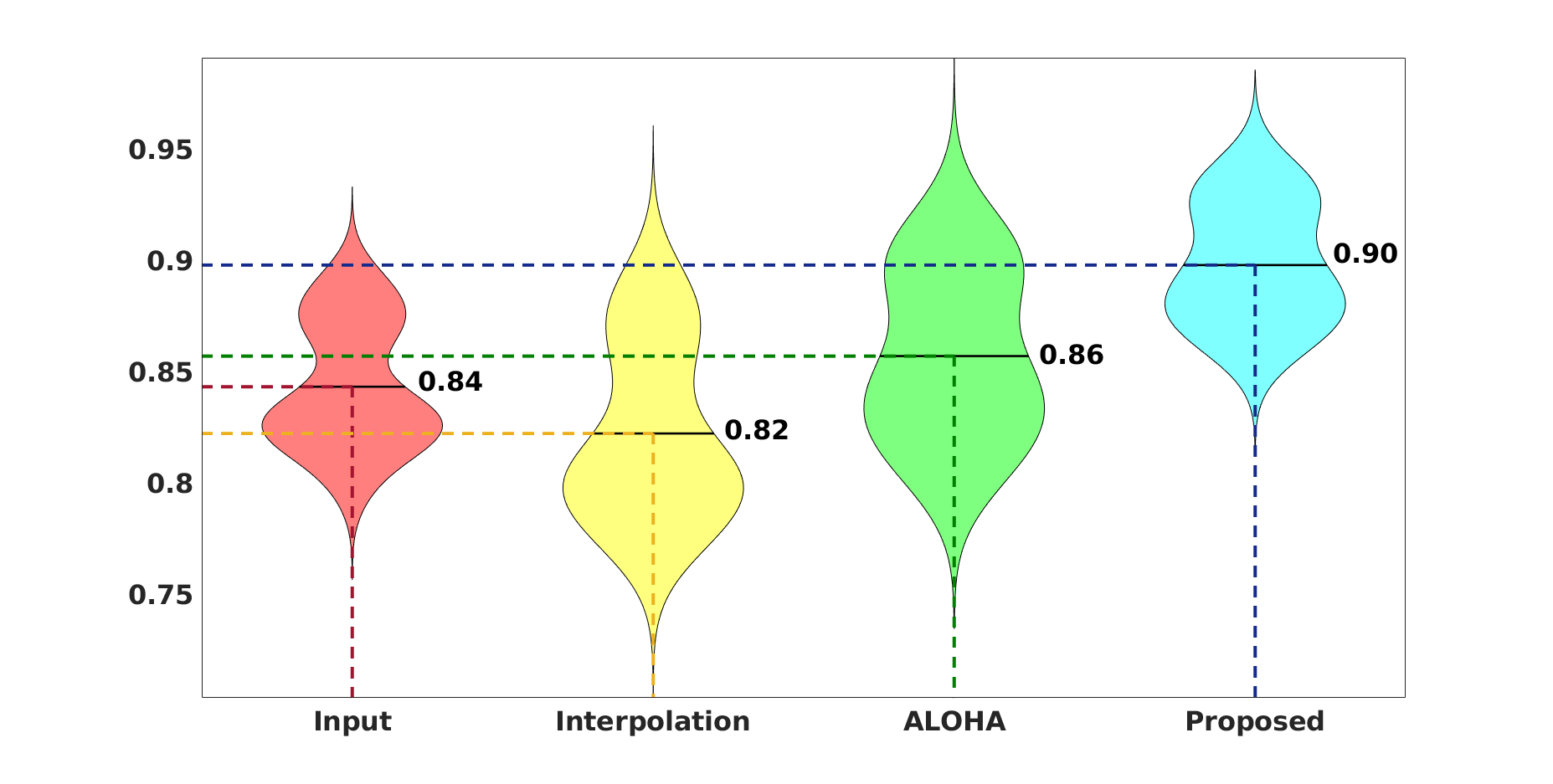}\hspace*{-0.5cm}\includegraphics[width=7.cm]{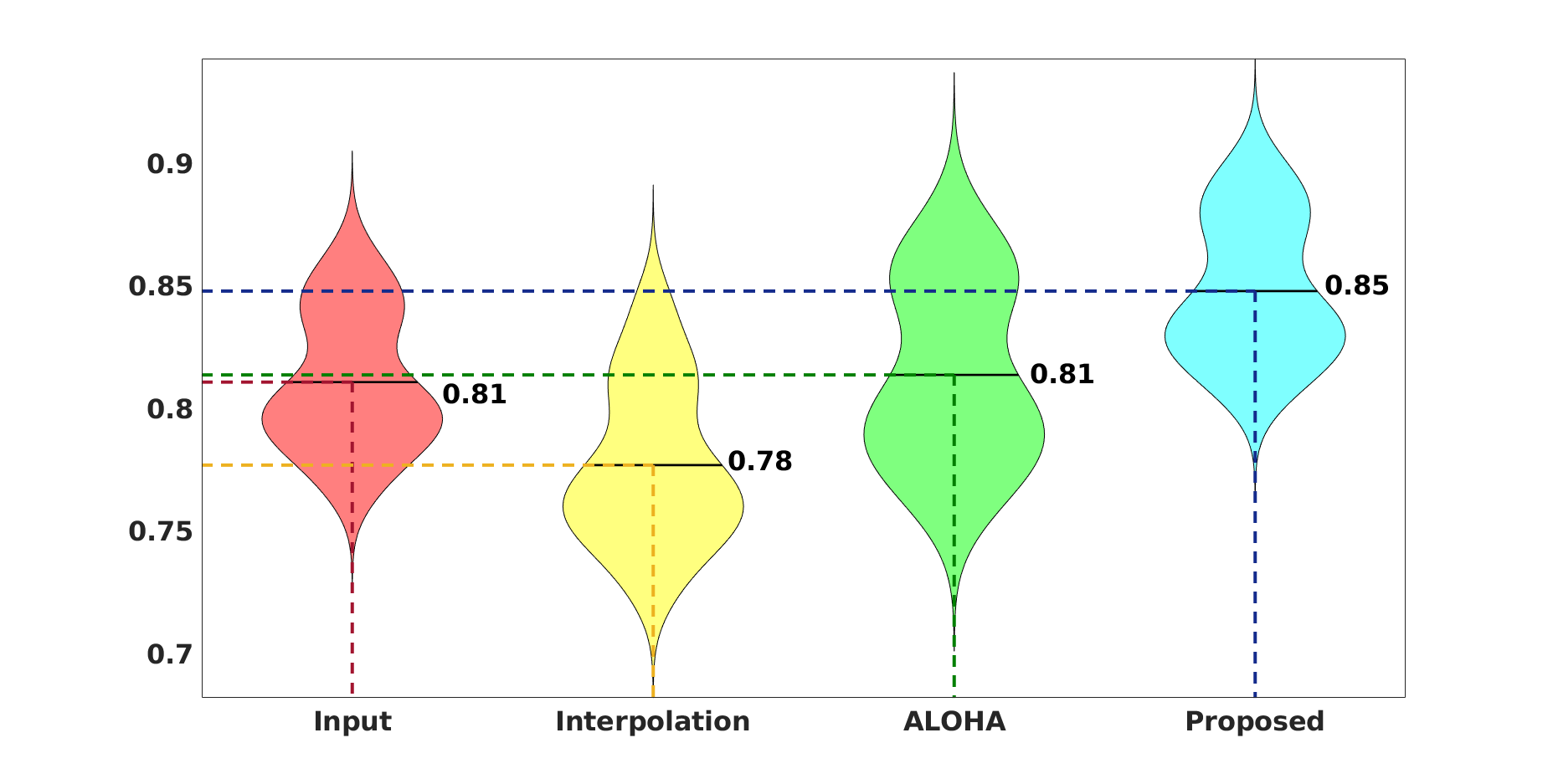}\hspace*{-0.5cm}\includegraphics[width=7.cm]{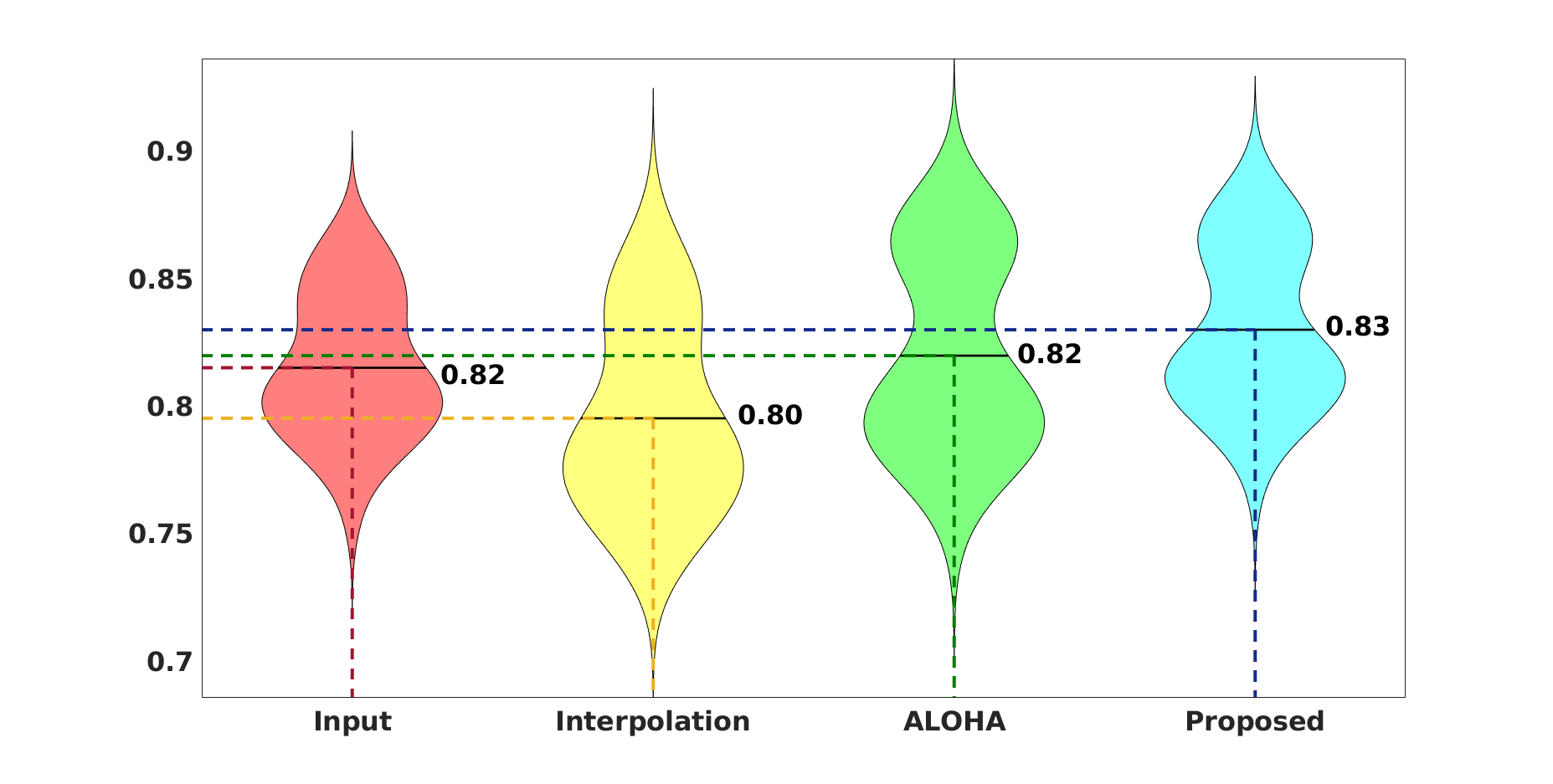}}
		\centerline{\mbox{(c) SSIM value distribution }}
	\caption{Abdominal region convex probe B-mode Reconstruction CNR, PSNR and SSIM value distribution of 100 images from various RF sub-sampling scheme: (First column) x4 Rx sub-sampling,
	(middle column) x8 Rx sub-sampling, (last column) 4x2 Rx-Xmit sub-sampling.}
	\label{fig:results_CNR_PSNR_SSIM_Con}
\end{figure*}

We compared the CNR, PSNR, and SSIM distributions for reconstructed B-mode images obtained from 100 test frames of a convex array.  As shown in Fig.~\ref{fig:results_CNR_PSNR_SSIM_Con}(a), compared to the linear interpolation and ALOHA based interpolation, the proposed method achieved CNR values of $1.75$, $1.70$ and $1.49$ in $\times4$ Rx, $\times8$ Rx and $4\times2$ Rx-Xmit sampling schemes, respectively.  These values are $24.11\%$, $23.19\%$ and $4.20\%$ higher than those for sub-sampled input, $19.05\%$, $25.93\%$ and $7.19\%$ higher than those for linear interpolation, and $12.18\%$, $15.65\%$ and $14.62\%$ higher than those for ALOHA.

 In this experiment, the quantitative improvement of contrast was high as in the linear array cases, and it is remarkable that accurate reconstruction was still obtained for the abdominal region, which was never observed with the network trained using the linear array transducer data. The results confirmed the generalization power of the algorithm.

Fig.~\ref{fig:results_CNR_PSNR_SSIM_Con}(b) compares the PSNR distribution in $4\times$ Rx, $8\times$ Rx, and $4\times 2$ Rx-Xmit sub-sampling schemes.  Compared to linear interpolation, the proposed deep learning method showed $3.22$dB, $2.69$dB, and $1.26$dB improvement on average for $4\times$ Rx, $8\times$ Rx, and $4\times 2$ Rx-Xmit sub-sampling schemes, respectively.  In comparison to ALOHA, the proposed deep learning method showed $0.98$dB, $1.09$dB, and $1.05$dB improvement on average for $4\times$ Rx, $8\times$ Rx, and $4\times 2$ Rx-Xmit sub-sampling schemes, respectively. 

In the convex array, the structural similarity (SSIM) of the DAS beam-former images in the proposed algorithm was  higher than that obtained with contemporary methods.  Fig.~\ref{fig:results_CNR_PSNR_SSIM_Con}(c) compares the SSIM in $4\times$ Rx, $8\times$ Rx, and $4\times 2$ Rx-Xmit sub-sampling schemes. Compared to linear interpolation, the proposed deep learning method showed $9.76\%$, $8.97\%$, and $3.75\%$ improvement in $4\times$ Rx, $8\times$ Rx, and $4\times 2$ Rx-Xmit sub-sampling schemes, respectively.  In comparison to ALOHA, the proposed deep learning method showed $4.65\%$, $4.94\%$, and $1.22\%$ improvement in $4\times$ Rx, $8\times$ Rx, and $4\times 2$ Rx-Xmit sub-sampling schemes, respectively.

\section{Conclusions}
\label{sec:conclusion}

In this paper, we presented a novel deep learning approach for accelerated B-mode ultrasound imaging. Inspired by the recent discovery of a close link between deep neural network and Hankel matrix decomposition, we searched for a signal domain in which the Hankel structured matrix is sufficiently low-ranked.  Our analysis showed that there are significant redundancies in the Rx-Xmit and Rx-SL domains, which results in a low-rank Hankel matrix. Thus, to exploit the redundancy in the RF domain, the proposed CNN was applied to the Rx-Xmit or Rx-SL domains. In contrast to existing CS approaches that require hardware changes or computationally expensive algorithms, the proposed method does not require any hardware change and can be applied to any B-mode ultrasound system or transducer.  Moreover, thanks to the exponentially increasing expressiveness of deep networks, PSNR, SSIM, and CNR were significantly improved over ALOHA and other existing methods, and the run-time complexity was orders of magnitude faster. Therefore, this method can be an important platform for RF sub-sampled US imaging.

\section{Acknowledgement}

This work was supported by the National Research Foundation of Korea, Grant number NRF-2016R1A2B3008104.

\section*{Appendix A: Complexity Analysis}

According to the analysis in \cite{jin2016general}, the complexity of ALOHA is mainly determined by matrix inversion during the ADMM step. For a given $n_1\times n_2$ RF data, in our ALOHA implementation, 
 the annihilating filter size was $d_1\times d_2=7\times 7$, and the number of
adjacent frames was $N=10$. Thus, the computational complexity for each matrix inversion is
given by $\mathcal{O}(n_1n_2(d_1d_2N)^2+(d_1d_2N)^3)$ per iteration, which is equal to $\mathcal{O}(n_1n_2 470^2 +470^3)$, where $\mathcal{O}$ denotes the ``big O'' notation.
Furthermore, there are two matrix inversions for each ADMM step and at least 50 iterations were required for convergence.

On the other hand, in our CNN implementation, we used $d\times d=3\times 3$ filters and the number of channels was $K=64$. 
Since there exist no matrix inversions and all the operations were by convolutions, the computational complexity for each multi-channel convolution layer with 64 input and 64 output channels is given by $\mathcal{O}(n_1n_2  (dK)^2)=\mathcal{O}(n_1n_2  192^2)$.
In addition, the number of convolutional layers was 28.
Thus, the overall computational complexity of CNN is lower than that of ALOHA.


\end{document}